\documentclass[10pt,twocolumn,letterpaper]{article}

\usepackage[pagenumbers]{configuration/cvpr} %
\usepackage{bbding}

\newcommand\blfootnote[1]{%
  \begingroup
  \renewcommand\thefootnote{}\footnote{#1}%
  \addtocounter{footnote}{-1}%
  \endgroup
}

\usepackage[pagebackref,hyperindex,breaklinks,colorlinks,citecolor=blue]{hyperref} %
\usepackage[utf8]{inputenc}         %
\usepackage[T1]{fontenc}            %
\usepackage{url}                    %
\usepackage{booktabs}               %
\usepackage{amsfonts}               %
\usepackage{nicefrac}               %
\usepackage{microtype}              %
\usepackage[dvipsnames]{xcolor}     %

\usepackage[utf8]{inputenc}         %
\usepackage[T1]{fontenc}            %
\usepackage{url}                    %
\usepackage{booktabs}               %
\usepackage{amsfonts}               %
\usepackage{nicefrac}               %
\usepackage{microtype}              %
\usepackage{xcolor}                 %

\usepackage[flushleft]{threeparttable}
\usepackage{float}
\usepackage{graphicx}
\usepackage{multirow}
\usepackage{xspace}
\usepackage{natbib}
\usepackage{enumitem}
\usepackage[font=small]{caption}
\usepackage{autobreak}
\usepackage{sidecap}
\usepackage{wrapfig}
\usepackage[toc, page, header]{appendix}
\newcommand{\algopt}{RDED\xspace}
\newcommand{\IPC}{\texttt{IPC}\xspace}
\definecolor{purple}{rgb}{0.74,0,0.74}
\definecolor{lemon}{rgb}{0.74,0.74,0}
\definecolor{turquoise}{rgb}{0,0.74,0.74}
\definecolor{emerald}{rgb}{0,0.5,0}

\definecolor{cvprblue}{rgb}{0.21,0.49,0.74}

\usepackage{amsmath,amsthm,amssymb}

\newtheorem{definition}{Definition}
\newtheorem{proposition}{Proposition}

\providecommand{\abs}[1]{\left\lvert#1\right\rvert}

\providecommand{\R}{\mathbb{R}} %

\providecommand{\E}{{\mathbb E}}
\providecommand{\E}[1]{{\mathbb E}\left.#1\right. }        %
\providecommand{\EEb}[2]{{\mathbb E}_{#1}\left[#2\right] } %

\DeclareMathOperator*{\argmin}{arg\,min\,}
\DeclareMathOperator*{\argmax}{arg\,max\,}

\let\lll\ll
\renewcommand{\ll}{\mathbf{l}}

\providecommand{\tt}{\mathbf{t}}

\providecommand{\xx}{\mathbf{x}}

\providecommand{\mtheta}{\boldsymbol{\theta}}

\providecommand{\cA}{\mathcal{A}}

\providecommand{\cH}{\mathcal{H}}

\providecommand{\cL}{\mathcal{L}}

\providecommand{\cP}{\mathcal{P}}
\providecommand{\cQ}{\mathcal{Q}}

\providecommand{\cS}{\mathcal{S}}
\providecommand{\cT}{\mathcal{T}}
\providecommand{\cU}{\mathcal{U}}
\providecommand{\cV}{\mathcal{V}}
\providecommand{\cX}{\mathcal{X}}
\providecommand{\cY}{\mathcal{Y}}

\newenvironment{talign*}
{\csname align*\endcsname}
{\endalign}

\usepackage{algorithm}
\usepackage[noend]{algpseudocode}

\algnewcommand\algorithmicforeach{\textbf{for each}}
\algdef{S}[FOR]{ForEach}[1]{\algorithmicforeach\ #1\ \algorithmicdo}

\errorcontextlines\maxdimen

\makeatletter
\newcommand*{\algrule}[1][\algorithmicindent]{\makebox[#1][l]{\hspace*{.5em}\thealgruleextra\vrule height \thealgruleheight depth \thealgruledepth}}%
\newcommand*{\thealgruleextra}{}
\newcommand*{\thealgruleheight}{.75\baselineskip}
\newcommand*{\thealgruledepth}{.25\baselineskip}

\newcount\ALG@printindent@tempcnta
\def\ALG@printindent{%
	\ifnum \theALG@nested>0%
	\ifx\ALG@text\ALG@x@notext%
	\else
		\unskip
		\addvspace{-1pt}%
		\ALG@printindent@tempcnta=1
		\loop
		\algrule[\csname ALG@ind@\the\ALG@printindent@tempcnta\endcsname]%
		\advance \ALG@printindent@tempcnta 1
		\ifnum \ALG@printindent@tempcnta<\numexpr\theALG@nested+1\relax%
			\repeat
		\fi
	\fi
}%
\usepackage{etoolbox}
\patchcmd{\ALG@doentity}{\noindent\hskip\ALG@tlm}{\ALG@printindent}{}{\errmessage{failed to patch}}
\makeatother

\newbox\statebox
\newcommand{\myState}[1]{%
	\setbox\statebox=\vbox{#1}%
	\edef\thealgruleheight{\dimexpr \the\ht\statebox+1pt\relax}%
	\edef\thealgruledepth{\dimexpr \the\dp\statebox+1pt\relax}%
	\ifdim\thealgruleheight<.75\baselineskip
		\def\thealgruleheight{\dimexpr .75\baselineskip+1pt\relax}%
	\fi
	\ifdim\thealgruledepth<.25\baselineskip
		\def\thealgruledepth{\dimexpr .25\baselineskip+1pt\relax}%
	\fi
	\State #1%
	\def\thealgruleheight{\dimexpr .75\baselineskip+1pt\relax}%
	\def\thealgruledepth{\dimexpr .25\baselineskip+1pt\relax}%
}

\def\paperID{16198} %
\def\confName{CVPR}
\def\confYear{2024}

\title{On the Diversity and Realism of Distilled Dataset: \\ An Efficient Dataset Distillation Paradigm}

\author{Peng Sun$^{2,1}$ \quad Bei Shi$^{1}$ \quad Daiwei Yu$^{3,*}$ \quad Tao Lin$^{1,\dagger} $ \\
$^{1}$Westlake University \quad $^{2}$Zhejiang University \quad $^{3}$ Independent Researcher \\
{\tt\small sunpeng@westlake.edu.cn, shibei0430@gmail.com, ydw.ccm@gmail.com, lintao@westlake.edu.cn} \\
}

\begin{document}

\twocolumn[{%
            \renewcommand\twocolumn[1][]{#1}%
            \maketitle
            \begin{center}
                \centering
                \includegraphics[width=17.5cm]{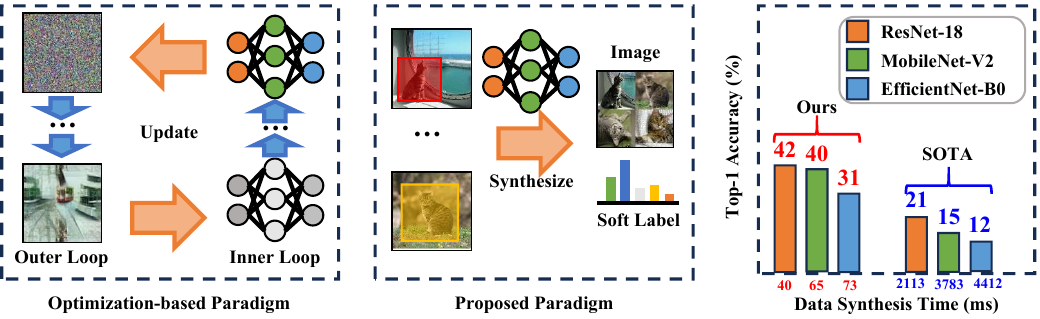}
                \vspace{-2em}
                \captionof{figure}{
                    \small
                    \textbf{Proposed paradigm vs.\ optimization-based paradigm.}
                    Left is the mainstream optimization-based dataset distillation and middle is our proposed non-optimizing paradigm.
                    Right is top-1 validation accuracy vs.\ synthesis time per image on ImageNet-1K with $\IPC = 10$ (10 Images Per Class).
                    Models used for distillation include ResNet-18, EfficientNet-B0, and MobileNet-V2; we use ResNet-18 for evaluation.
                    \looseness=-1
                } \label{fig:contribution}
            \end{center}
        }]

\begin{abstract}
    \vspace{-10pt}
    \blfootnote{$^\dagger$ Corresponding author.}
    \blfootnote{$*$ Work done during Daiwei's visit to Westlake University.}
    Contemporary machine learning, which involves training large neural networks on massive datasets, faces significant computational challenges.
    Dataset distillation, as a recent emerging strategy, aims to compress real-world datasets for efficient training.
    However, this line of research currently struggles with large-scale and high-resolution datasets, hindering its practicality and feasibility.
    Thus, we re-examine existing methods and identify three properties essential for real-world applications: realism, diversity, and efficiency.
    As a remedy, we propose \algopt, a novel computationally-efficient yet effective data distillation paradigm, to enable both diversity and realism of the distilled data.
    Extensive empirical results over various model architectures and datasets demonstrate the advancement of \algopt: we can distill a dataset to $10$ images per class from full ImageNet-1K~\cite{deng2009imagenet} within $7$ minutes, achieving a notable $42\%$ accuracy with ResNet-18 \cite{he2016deep} on a single RTX-4090 GPU (while the SOTA only achieves $21\%$ but requires $6$ hours).
    Code: \url{https://github.com/LINs-lab/RDED}.
\end{abstract}

\setlength{\parskip}{0pt plus0pt minus0pt}

\section{Introduction}
\label{sec:intro}

The success of modern deep learning could be largely attributed to the fact of scaling and increasing both neural architectures and training datasets~\cite{he2016deep,ioffe2015batch,goodfellow2014generative,kaplan2020scaling}.
Though this pattern shows great potential to propel artificial intelligence forward, the challenge of high computational requirements remains a noteworthy concern \cite{yu2023dataset,cui2022dc,wang2018dataset}.
Dataset distillation methods have recently emerged \cite{yu2023dataset, wang2018dataset} and attracted attention for their exceptional performance \cite{cazenavette2022dataset, cazenavette2023generalizing, zhao2020dataset, wang2022cafe, zhao2023dataset, zhao2023improved}.
The key idea is compressing original full datasets by synthesizing and optimizing a small dataset, where training a model using the synthetic dataset can achieve a similar performance to the original.

However, these methods suffer a high computational burden \cite{cui2023scaling,yin2023squeeze} due to the bi-level optimization-based paradigm.
Moreover, the synthetic images exhibit certain \emph{non-realistic} features (see Figure~\ref{fig:fruits_mtt} and \ref{fig:fruits_sre2l}) that have materialized due to overfitting to a specific architecture used during the optimization process, which leads to difficulties in generalizing to other architectures~\cite{cazenavette2023generalizing,shao2023generalized}.

A notable work \cite{cazenavette2023generalizing} investigates the relationship between realism and expressiveness in synthetic datasets. The findings reveal a trade-off: more realistic images come at the sacrifice of expressiveness. While realism aids in generalizing across different architectures, it hurts distillation performance. Conversely, prioritizing expressiveness over realism can enhance distillation performance but may impede cross-architecture generalization.

Inspired by these insights, we introduce an \underline{Realistic}, \underline{Diverse}, and \underline{Efficient} Dataset \underline{Distillation} (RDED) method.
Our goal is to achieve diversity (expressiveness) and realism simultaneously across varying datasets, ranging from CIFAR-10 to ImageNet-1K.
Specifically, we directly crop and select realistic patches from the original data to maintain realism. To ensure the greatest possible diversity, we stitch the selected patches into the new images as the synthetic dataset. It is noteworthy that our method is non-optimization-based, so it can also achieve high efficiency, making it well-suited for processing large-scale, high-resolution datasets.

The key contributions of this work can be summarized as:
\begin{itemize}[nosep, leftmargin=12pt]
    \item We first investigate the limitations of existing dataset distillation methods and define three key properties for effective dataset distillation on large-scale high-resolution datasets: realism, diversity, and efficiency.
    \item We introduce the definitions of diversity ratio and realism score backed by $\cV$-information theory \cite{xu2020theory}, together with an optimization-free efficient paradigm, to enhance diversity and realism of the distilled data.
    \item Extensive experiments demonstrate the effectiveness of our method: it not only achieves a top-1 validation accuracy that is twice the current SOTA---SRe$^2$L~\citep{yin2023squeeze}, but it also operates at a speed $52$ times faster (see Figure~\ref{fig:contribution}).
\end{itemize}

\section{Related Work}
\label{sec:relatedwork}

Dataset distillation, as proposed by \citet{wang2018dataset}, condenses large datasets into smaller ones without sacrificing performance.
These methods fall into four main categories.

\paragraph{Bi-level optimization-based distillation.}
A line of work seeks to minimize the surrogate models learned from both synthetic and original datasets, depending on their metrics, namely, the matching gradients \cite{zhao2020dataset,kim2022dataset,zhang2023accelerating}, features \cite{wang2022cafe}, distribution \cite{zhao2023dataset,zhao2023improved}, and training trajectories \cite{cazenavette2022dataset,cui2022dc,du2023minimizing,cui2023scaling,yu2023dataset,guo2023towards}.
Notably, trajectory matching-based techniques have demonstrated remarkable performance across various benchmarks with low \IPC.
However, the synthetic data often overfit to a specific model architecture, struggling to generalize to others.

\paragraph{Distillation with prior regularization.}

\citet{cazenavette2023generalizing} suggest that direct pixel space parameterization is a key factor for the architecture transferability issue, and propose GLaD to integrate a generative prior for dataset distillation to enhance generalization across any distillation method.
However, bi-level optimization-based methods, especially those that entail prior regularization, face computational challenges and memory issues \cite{cui2023scaling}.

\paragraph{Uni-level optimization-based distillation.}
Kernel ridge-regression methods~\cite{zhou2022dataset,loo2022efficient}, with uni-level optimization,
effectively reduce training costs \cite{zhou2022dataset} and enhancing performance \cite{loo2022efficient}.
However, due to the resource-intensive nature of inverting matrix operations, scaling these methods to larger \IPC remains challenging.
Unlike NTK-based solutions, \citet{yin2023squeeze} propose to decouple the bi-level optimization of dataset condensation into two single-level learning procedures, resulting in a more efficient framework.

\paragraph{CoreSet selection-based distillation.}
CoreSet selection, akin to traditional dataset distillation, focuses on identifying representative samples using provided images and labels.
Various difficulty-based metrics are proposed to assess the sample importance, e.g., the forgetting score \cite{toneva2018empirical}, memorization \cite{feldman2020neural}, EL2N score \cite{paul2021deep}, diverse ensembles \cite{meding2021trivial}.

\section{On the Limits of Dataset Distillation}
\begin{figure*}
    \centering
    \begin{subfigure}{0.31\linewidth}
        \includegraphics[width=5.27cm]{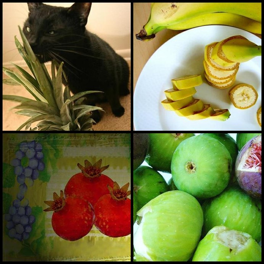}
        \caption{Random selection of original dataset}
        \label{fig:fruits_random}
    \end{subfigure}
    \hfill
    \begin{subfigure}{0.31\linewidth}
        \includegraphics[width=5.27cm]{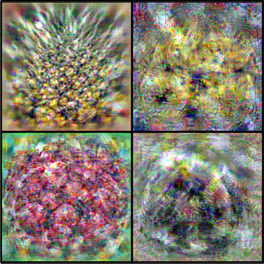}
        \caption{MTT~\cite{cazenavette2022dataset}}
        \label{fig:fruits_mtt}
    \end{subfigure}
    \hfill
    \begin{subfigure}{0.31\linewidth}
        \includegraphics[width=5.27cm]{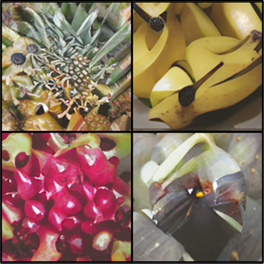}
        \caption{GLaD~\cite{cazenavette2023generalizing}}
        \label{fig:fruits_glad}
    \end{subfigure}
    \vfill
    \begin{subfigure}{0.31\linewidth}
        \includegraphics[width=5.27cm]{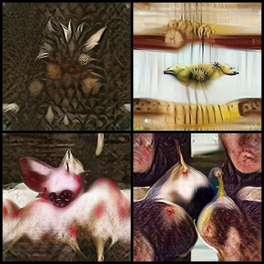}
        \caption{SRe$^2$L~\cite{yin2023squeeze}}
        \label{fig:fruits_sre2l}
    \end{subfigure}
    \hspace{12pt}
    \begin{subfigure}{0.31\linewidth}
        \includegraphics[width=5.27cm]{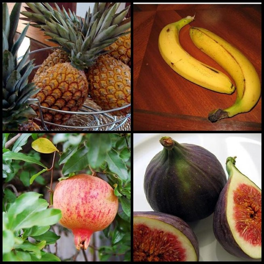}
        \caption{Herding~\cite{welling2009herding}}
        \label{fig:fruits_herding}
    \end{subfigure}
    \hspace{12pt}
    \begin{subfigure}{0.31\linewidth}
        \includegraphics[width=5.27cm]{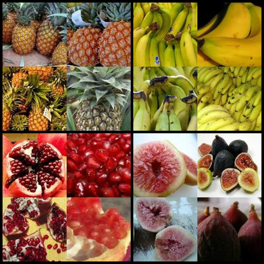}
        \caption{\algopt (Ours)}
        \label{fig:fruits_ours}
    \end{subfigure}
    \vspace{-2.em}
    \caption{
        \small
        \textbf{Visualization of images synthesized using various dataset distillation methods}.
        We consider the ImageNet-Fruits \cite{cazenavette2022dataset} dataset, comprising a total of 10 distinct fruit types, with a resolution of $128 \times 128$.
        There are four specific classes for each method, namely, 1) \textit{Pineapple}, 2) \textit{Banana}, 3) \textit{Pomegranate}, and 4) \textit{Fig}.
        Note that MTT~\cite{cazenavette2022dataset}, GLaD~\cite{cazenavette2023generalizing}, SRe$^2$L~\cite{yin2023squeeze}, and Herding~\cite{welling2009herding}, are four representative methods of conventional dataset distillation paradigms discussed in Section~\ref{sec:relatedwork} and Section~\ref{sec:pitfalls} (see Appendix~\ref{sec:aux_visualization} for more visualization).
        In general, ensuring both superior \emph{realism} and \emph{diversity} simultaneously is challenging for methods other than ours and GLaD.
    }
    \vspace{-1.5em}
    \label{fig:comparison}
\end{figure*}
We start by clearly defining the concept of dataset distillation and then reveal the primary challenges in this field.

\subsection{Preliminary}

The goal of dataset distillation is to synthesize a smaller distilled dataset, denoted as $\cS = (X,Y) = \{ \xx_{j}, y_{j} \}_{j=1}^{\abs{\cS}}$, that captures the essential characteristics of a larger dataset $\cT= (\hat{X},\hat{Y}) =\{ \hat{\xx}_{i}, \hat{y}_{i} \}_{i=1}^{\abs{\cT}}$.
Here, the distilled dataset $\cS$ is generated by an algorithm $\cA$ such that $\cS \in \cA(\cT)$, where the size of $\cS$ is considerably smaller than $\cT$ (i.e., $\abs{\cS} \lll \abs{\cT}$).
Each ${y}_j \in Y$ corresponds to the synthetic distilled label for the sample ${\xx}_j \in X$, and a similar definition can be applied to $(\hat{\xx}_{i}\in\hat{X},\hat{y}_{i}\in\hat{Y})$.
The key motivation for dataset distillation is to create a dataset $\cS$ that allows models to achieve performance within an acceptable deviation $\epsilon$ from those trained on $\cT$. Formally, this is expressed as:
\begin{equation}
    \sup \left\{ | \ell(\phi_{\mtheta_{\cT}}(\xx), y) - \ell(\phi_{\mtheta_{\cS}}(\xx), y) | \right\}_{(\xx, y) \sim \cT} \leq \epsilon, \label{eq:ddobj}
\end{equation}
where $\mtheta_{\cT}$ is the parameter set of the neural network $\phi$ optimized on $\cT$:
\begin{equation}
    \mtheta_{\cT} = \argmin_{\mtheta} \mathbb{E}_{(\xx, y) \in \cT} \left[ \ell(\phi_{\mtheta}(\xx), y) \right],
\end{equation}
with $\ell$ representing the loss function. A similar definition applies to $\mtheta_{\cS}$.

\newcommand{\GreenCMB}{\textcolor{green}{\CheckmarkBold}}
\begin{table}[t]
    \centering
    \setlength\tabcolsep{2pt}
    \scalebox{0.85}{
        \begin{tabular}{@{}l|ccc|cc@{}}
            \toprule[2pt]
                     & \multicolumn{3}{c|}{Property} & \multicolumn{2}{c}{Dataset}                                                 \\ \midrule
            Method   & Diversity                     & Realism                     & Efficiency   & Large-scale  & High-resolution \\
            MTT      & \GreenCMB                     & \XSolidBrush                & \XSolidBrush & \XSolidBrush & \XSolidBrush    \\
            GLaD     & \GreenCMB                     & \GreenCMB                   & \XSolidBrush & \XSolidBrush & \GreenCMB       \\
            SRe$^2$L & \Checkmark                    & \Checkmark                  & \Checkmark   & \Checkmark   & \Checkmark      \\
            Herding  & \Checkmark                    & \Checkmark                  & \GreenCMB    & \Checkmark   & \Checkmark      \\
            Ours     & \GreenCMB                     & \GreenCMB                   & \GreenCMB    & \GreenCMB    & \GreenCMB       \\ \bottomrule[2pt]
        \end{tabular}
    }
    \caption{
        \small
        \textbf{Properties and performance of various representative SOTA dataset distillation methods.}
        We give a summary of the properties of different methods and their performance on large-scale or high-resolution datasets, where \GreenCMB, \Checkmark, and \XSolidBrush, denote ``Superior'', ``Satisfactory'', and ``Bad'' respectively.
    }
    \vspace{-1.75em}
    \label{tab:limitation}
\end{table}

\paragraph{The properties of optimal dataset distillation.}
The effectiveness and utility of dataset distillation methods rely on key properties outlined in Definition~\ref{def:properties}. These properties are crucial for creating datasets efficiently, which in turn, enhances model training and generalization.

\begin{definition}[Properties of distilled data] \label{def:properties}
    Consider a family of observer models $\cV$\footnote{
    $\cV$ includes different observer models, for instance, humans $\phi_{\mathrm{h}}$ and pre-trained models $\phi_{\mtheta_{\cT}}$.
    Here, $\phi_{\mathrm{h}}$ is an abstraction representing human predictive behavior.
    }.
    The core attributes of a distilled dataset $\cS=(X,Y) \in \cA(\cT)$ are defined as follows:
    \begin{enumerate}
        \item \textbf{Diversity:} Essential for robust learning and generalization, a high-quality dataset should cover a wide range of samples $X$ and labels $Y$ \cite{sorscher2022beyond, kaplan2020scaling,radford2021learning}. This ensures exposure to diverse features and contexts.
        \item \textbf{Realism:} Critical for cross-architecture generalization, realistic distilled samples $X$ and labels $Y$ should be accurately predicted and matched by various observer models from $\cV$. It is important to avoid features or annotations that are overly tailored to a specific model \cite{cazenavette2022dataset,zhao2020dataset, zhao2023dataset}.
        \item \textbf{Efficiency:} A determinant for the feasibility of dataset distillation, addressing the computational and memory challenges is crucial for scaling the distillation algorithm $\cA$ to large datasets \cite{cui2023scaling,yin2023squeeze}. \looseness=-1
    \end{enumerate}
\end{definition}

\subsection{Pitfalls of Conventional Dataset Distillation} \label{sec:pitfalls}
In response to the properties of the optimal dataset distillation, in this section, we conduct a comprehensive examination of four conventional dataset distillation paradigms discussed in Section~\ref{sec:relatedwork}.
Limitations are detailed below and summarized in Table~\ref{tab:limitation} (see more details in Appendix~\ref{sec:aux_visualization}).

\begin{itemize}
    \item \textbf{Bi-level optimization-based distillation.}
          Conventional dataset distillation methods \cite{cazenavette2022dataset, zhao2020dataset, zhao2023dataset} suffer from noise-like \emph{non-realistic} patterns (see Figure \ref{fig:fruits_mtt}) in distilled high-resolution images and overfit the specific architecture used in training \cite{cazenavette2023generalizing}, which hurt its cross-architecture generalization ability \cite{cazenavette2023generalizing}.
          However, these methods suffer a \emph{high computational burden} \cite{cui2023scaling,yin2023squeeze} due to the bi-level optimization-based paradigm.

    \item \textbf{Distillation with prior regularization.}
          \citet{cazenavette2023generalizing} identify the source of the architecture overfitting issue, and thus enhances the realism (see Figure \ref{fig:fruits_glad}) of synthetic images and the cross-architecture generalization.
          The current remedy inherits the \emph{low efficiency} of bi-level optimization-based distillation, and thus still cannot generalize to large-scale datasets.
    \item \textbf{Uni-level optimization-based distillation.}
          As a remedy for the former research, \citet{yin2023squeeze}---as the latest progress in the field---alleviate the \emph{efficiency} and \emph{realism} challenges (see Figure \ref{fig:fruits_sre2l}) and propose SRe$^2$L to distill large-scale, high-resolution datasets, e.g., ImageNet-1K.

          Yet, SRe$^2$L is hampered by a limited \emph{diversity} problem arising from its synthesis approach, which involves extracting knowledge from a pre-trained model containing only partial information of the original dataset \cite{yin2020dreaming}.

    \item \textbf{CoreSet selection-based distillation.}
          CoreSet selection methods \cite{sorscher2022beyond,tan2023data,welling2009herding} serve to \emph{efficiently} distill datasets by isolating a CoreSet containing \emph{realistic} images (see Figure \ref{fig:fruits_herding}).
          However, the advances come at the cost of limited information representation (\emph{data diversity})~\cite{paul2021deep}, leading to a catastrophically degraded performance \cite{tan2023data}.
\end{itemize}

\begin{figure*}[ht]
    \centering
    \includegraphics[width=\linewidth]{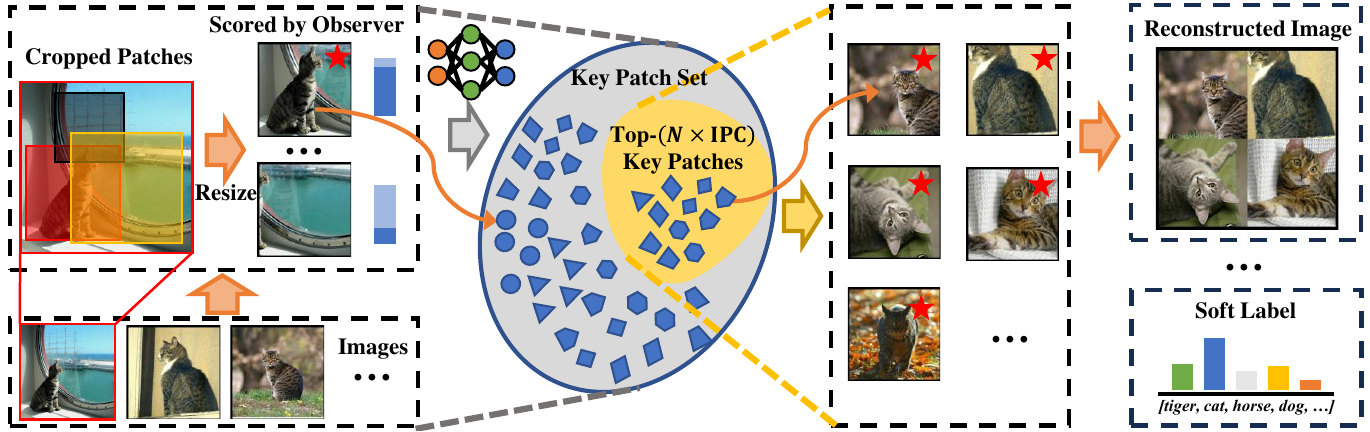}
    \vspace{-2em}
    \caption{
        \small
        \textbf{Visualization of our proposed two-stage dataset distillation framework.}
        Stage 1: We crop each original image into several patches and rank them using the realism scores calculated by the observer model. Then, we choose the top-1-scored patch as the key patch.
        For the key patches within a class, we re-select the top-$N \times \IPC$ patches based on their scores, where $N = 4$ in this case.
        Stage 2: We consolidate every $N$ selected patches from Stage 1 into a single new image that shares the same resolution with each original image, resulting in $\IPC$-numbered distilled images per class.
        These images are then relabeled using the pre-trained observer model.
    }
    \vspace{-1em}
    \label{fig:framework}
\end{figure*}

\section{Methodology}
To tackle the remaining concern of distilling high-resolution and large-scale image datasets, in this section, we articulate an novel unified dataset distillation paradigm---\algopt---that prioritizes both \emph{diversity} and \emph{realism} within the distilled dataset, yet being \emph{efficient}.
\looseness=-1

\subsection{Enhancing Data Diversity and Realism}

\paragraph{Establishing a $\cV$-information-based objective for distilled data.}
Drawing on the artificial intelligence learning principles of \textit{parsimony} and \textit{self-consistency} from \cite{ma2022principles}, we strive to ensure that the models trained on the distilled dataset embody these principles.
To achieve this, we aim to construct a representation $Y$ of the input data $X$ that is structured (\textit{parsimony}) and rich in information (\textit{self-consistency}).
Consequently, we reinterpret the objective of dataset distillation in~\eqref{eq:ddobj}, as the structured and sufficient information inherent in the original full dataset $\cT$:
\begin{equation}
    \textstyle
    \cS = \underset{(X,Y)\in\cA(\cT)}{\argmax} {I_{\cV}(X \rightarrow Y)} \,, \label{eq:principle}
\end{equation}
where $I_{\cV}$ denotes the predictive $\cV$-information \cite{xu2020theory} from $X$ to $Y$, which can be further defined as:
\begin{equation}
    \textstyle
    I_{\cV}(X \rightarrow Y) = \underbrace{H_{\cV}(Y|\varnothing)}_{\text{diversity}} - \underbrace{H_{\cV}(Y|X)}_{\text{realism}} \,, \label{eq:vinfo}
\end{equation}
where $H_{\cV}(Y|X)$ and $H_{\cV}(Y|\varnothing)$ denote, respectively, the predictive conditional $\cV$-entropy~\cite{xu2020theory} with observed side information $X$ and no side information $\varnothing$.

\paragraph{Explicitizing the diversity and realism via $\cV$-information.}
Building upon Definition~\ref{def:properties}, maximizing $H_{\cV}(Y|\varnothing)$ can enhance the uncertainty/diversity of representations $Y$ measured by the observer models in $\cV$.
Simultaneously, minimizing $H_{\cV}(Y|X)$ aims to improve the predictiveness/realism of the data pairs $(X, Y)$ \cite{xu2020theory,ethayarajh2022understanding}.
Therefore, the objective of \eqref{eq:principle} is equivalent to maximize the first term $H_{\cV}(Y|\varnothing)$ while minimizing the second term $H_{\cV}(Y|X)$ in~\eqref{eq:vinfo}, to achieve the improved data diversity and realism.

\paragraph{Approximating and maximizing $\cV$-information.}
For the sake of computational feasibility, we restrict ourselves to the case where the predictive family $\cV$ includes only humans and a single pre-trained observer model associated with dataset $\cT$, denoted as $\cV = \{\phi_{\mathrm{h}}, \phi_{\mtheta_{\cT}}\}$.
Given the computational challenges of solving \eqref{eq:principle} by maximizing both terms in \eqref{eq:vinfo} simultaneously, we decouple the terms in \eqref{eq:vinfo}, resulting in:
\begin{equation}
    \inf_{f \in \cV}
    \begin{cases}
        \EEb{ y \sim Y}{-\log{f[\varnothing](y)}} \\
        - \EEb{\xx, y \sim X, Y}{-\log{f[\xx](y)}}
    \end{cases} ,
    \label{eq:expvinfo}
\end{equation}
where $f[\xx]$ is probability measure on $Y$ based on the received information $\xx$, and $f[\xx](y) \in \R$ is the value of the density evaluated at $y \in Y$.
Then, we seek proxies to approximate the decoupled two terms in \eqref{eq:expvinfo} independently.

\begin{proposition}[Proxies on the diversity and realism of distilled data] \label{prop:vinfo_appr}
    Given a distilled dataset $\cS =(X,Y)$,
    we derive the following approximations to maximize the diversity term $H_{\mathcal{V}}(Y|\varnothing)$ and the realism term $-H_{\mathcal{V}}(Y|X)$:
    \begin{enumerate}
        \item The diversity ratio $H_{\cV}(Y|\varnothing) / H_{\cV}(\cT|\varnothing)$ is posited as a lower bound of the information preservation ratio from the original dataset $\mathcal{T}$ to the distilled one $\mathcal{S}$, justified by:
        \begin{equation}
            H_{\cV}(Y|\varnothing) \leq H_{\cV}(\cS|\varnothing) \leq H_{\cV}(\cT|\varnothing) \, .
        \end{equation}
        Therefore, we maximize diversity through preserving more information from the original dataset $\cT$.
        \item The realism score for a distilled sample $\xx$ and label $y$ from a pair $(\xx, y)$ is defined as:
        \begin{equation}
        -\ell(\phi_{\mtheta_{\cT}}(\xx), \phi_{\mathrm{h}}(\xx)) -\ell(\phi_{\mtheta_{\cT}}(\xx), y)\, .
        \end{equation}
        To enhance the realism score for each distilled pair $(\xx, y)$, we prioritize the distillation of sample $\xx$ with higher $-\ell(\phi_{\mtheta_{\mathcal{T}}}(\xx), \phi_{\text{h}}(\xx))$ and assign the label $y = \phi_{\mtheta_{\mathcal{T}}}(\xx)$.
    \end{enumerate}
\end{proposition}
\paragraph{Summary.}\emph{To bolster the two properties---diversity and realism---of our distilled dataset, we employ two practical proxies, namely 1) the diversity ratio, and 2) the realism score, as the approximation to design distillation algorithm $\cA$ (see Appendix~\ref{sec:aux_vinfo} for more (theoretical) analysis).}

\paragraph{Overview of our dataset distillation paradigm.}

To enhance the diversity and realism of our distilled dataset, we introduce a novel two-stage paradigm that practically utilizes the proposed two proxies in Proposition~\ref{prop:vinfo_appr} (See Figure~\ref{fig:framework} and Algorithm \ref{alg:framework}).
In particular, our objective is to preserve the information within a large number of sample pairs exhibiting high realism scores from the original full dataset $\cT$ into the distilled dataset $\mathcal{S}$.
This process unfolds in two stages:
\begin{itemize}
    \item First stage in Section \ref{sec:infoextract} \emph{extracts major information} (i.e., key sample pairs) with \emph{high realism score} from $\cT$.
    \item In the second stage (see Section \ref{sec:inforecon}), we aim to \emph{compress the extracted information} from the first stage into finite pixel space to form distilled images and \emph{relabel} them.
\end{itemize}

\subsection{Extracting Key Patches from Original Dataset} \label{sec:infoextract}
To extract the explicit key information from the original full dataset, we capture the key patches with high realism scores at the pixel space level and sample space level respectively.

\paragraph{Extracting key patch per image.}
Motivated by the common practice in Vision Transformer~\cite{dosovitskiy2021an,zagoruyko2016paying} that \emph{image patches} are sufficient to capture object-related information, we propose to learn the most realistic patch, $\xi_{i,\star}$, from a set of patches $\{ \xi_{i,k}\}$, which are extracted from a given image $\hat{\xx}_i \in \hat{X}$.
The whole procedure can be formulated as:
\begin{equation}
    \xi_{i,\star} = \argmax_{\xi_{i,k} \sim p( \xi_{i,k}| \hat{\xx}_i )}{ -\ell(\phi_{\mtheta_\cT}(\xi_{i,k}), \phi_{\mathrm{h}}(\xi_{i,k})) } \,,
\end{equation}
where the label $\phi_{\mathrm{h}}(\xi_{i,k})$ annotated by humans is given as $y_i$.
Therefore, $s_{i,k} := -\ell(\phi_{\mtheta_\cT}(\xi_{i,k}), y_i)$ represents the realism score for the patch $\xi_{i,k}$ and $s_{i,\star}$ denotes the highest one.

Let $\cT_c := \{ (\hat{\xx}, \hat{y}) \vert (\hat{\xx}, \hat{y}) \in \cT, \hat{y} \!=\! c \}$ denote a sub-dataset comprising all samples associated with class $c$ from the full dataset $\cT$.
Given the key patches and its corresponding scores $( \xi_{i,\star}, s_{i,\star} )$ for $\hat{\xx}_i \in \cT_c$, we form them into a set $\cQ_c$.

\paragraph{Capturing inner-class information.}
Solely relying on information extraction at the pixel space level
is inadequate in averting information redundancy at the sample space level.
To further extract key information from the original dataset, we consider a sample space level selection to further scrutiny of the selected patches from the previous stage.

More precisely, certain patches---denoted as $\cQ_c^\prime$---are selected based on a given pruning criteria $\bar{s}_{\star}$ defined over $\cQ_c$, aiming to capture the most impactful patches for class $c$, whose socres are larger than $\bar{s}_{\star}$.
This process is iteratively repeated for all classes of $\cT$.

\paragraph{Practical implementation.}
In practice, extracting all key patches from the entire $\cT_c$ and subsequently selecting the top patches based on scoring presents two significant challenges:
\begin{itemize}
    \item Iterating through each image in $\cT_c$ to identify crucial patches incurs a \emph{considerable computational overhead}.
    \item Utilizing a score-based selection strategy typically introduces \emph{distribution bias} within the chosen subset of the original dataset $\cT_c$, which hurts data diversity and adversely affects generalization (see Section~\ref{sec:ablationstudy} for more details).
\end{itemize}
To address the aforementioned issues, we propose the adoption of the random uniform data selection strategy\footnote{
    This treatment is motivated by the findings~\cite{coleman2019selection,tan2023data,sorscher2022beyond} on the impact of various data selection strategies, where random uniform selection is a \emph{lightweight} yet \emph{unbiased} data selection strategy.
} to derive a pre-selected subset $\cT_c^\prime \subset \cT_c$ (see settings in Section \ref{sec:expset}).
The subsequent inner-class information-capturing process is then performed exclusively on this subset $\cT_c^\prime$.

\subsection{Information Reconstruction of Patches} \label{sec:inforecon}
To effectively save the previously extracted key information in the limited pixel space and label space of the distilled dataset, we propose to reconstruct the information in patches.

\paragraph{Images reconstruction.}
The patch size is typically smaller than the dimensions of an expected distilled image, where directly utilizing the patches selected as distilled images may lead to sparse information in the pixel space.

Therefore, for a given class $c$ with a selected patch set $\cQ_c^\prime$, we randomly retrieve $N$ patches\footnote{
    We use $N$ and target resolution of distilled images to calculate and set the resolution of the patches, e.g., for an image with $224 \times 224$ resolution and set $N=4$, the resolution of patches is defined as $112 \times 112$.
    \label{fn:patchesnumber}
} without replacement to form a final image $\xx_j$ by applying the following operation:
\begin{equation} \label{eq:squeeze}
    \textstyle
    {\xx}_j = \text{concatenate}(\{{ \xi_{i,\star}}\}_{i=1}^N \subset \cQ_c^\prime) \,.
\end{equation}

\paragraph{Labels reconstruction.}
The previous investigation \cite{yun2021re} highlights a critical limitation associated with single-label annotations, wherein a random crop of an image may encompass an entirely different object than the ground truth, thereby introducing noisy or even erroneous supervision during the training process.
Consequently, relying solely on the simplistic one-hot label proves inadequate for representing an informative image, consequently constraining the effectiveness and efficiency of model learning \cite{yin2023squeeze}.

Inspired by this observation, we propose to re-label the squeezed multi-patches within the distilled images ${\xx}_j$, thereby encapsulating the informative label for the distilled images.
It can be achieved by employing the soft labelling approach \citep{shen2022fast} to generate region-level soft labels ${y}_{j,m} = \ell \left( \phi_{\mtheta_\cT} ( {\xx}_{j,m} ) \right)$, where ${\xx}_{j,m}$ is the $m$-th region in the distilled image and ${y}_{j,m}$ is the corresponding soft label.

\paragraph{Training with reconstructed labels.}
We train the student model $\phi_{\mtheta_\cS}$ on the distilled data using the following objective:
\begin{equation}
    \cL = - \sum_{j}{\sum_{m}{{y}_{j,m} \log{\phi_{\mtheta_\cS}( {\xx}_{j,m} )}}} \,.
\end{equation}

\begin{algorithm}[t]
    \caption{\algopt: An efficient framework for high-resolution dataset distillation (see Appendix~\ref{sec:imple_details} for more implementation details)}\label{alg:framework}
    \begin{algorithmic}
        \Statex \textbf{Input:} Original full dataset $\cT$, a corresponding pre-trained observer model $\phi_{\mtheta_{\cT}}$ and initial $\cS = \varnothing$.
        \For{$\cT_c^\prime \subset \cT_c \subset \cT$}
        \For{$(\hat{\xx_i}, \hat{y}_i) \in \cT_c^\prime$} \Comment{\textcolor{magenta}{Stage 1}}
        \State Crop $\hat{\xx_i}$ into $K$ patches $\{ \xi_{i,k} \}_{k=1}^K$
        \For{$k = 1$ to $K$}
        \State Calculate the score $s_{i,k} = -\ell({\phi_{\mtheta_{\cT}}}(\xi_{i,k}), \hat{y}_i)$
        \EndFor
        \State Select patch $\xi_{i,\star}$ from $ \{ \xi_{i,k} \}_{k=1}^K$ via $s_{i,\star}$
        \EndFor
        \State Select top-($N \times \IPC$) patches
        \For{$j=1$ to \IPC} \Comment{\textcolor{magenta}{Stage 2}}
        \State Squeeze $N$ selected patches into ${\xx}_j$
        \State Relabel ${\xx}_j$ with ${y}_j$
        \State $\cS=\cS \cup \{({\xx}_j, {y}_j)\}$
        \EndFor
        \EndFor
        \Statex \textbf{Output:} Small distilled dataset $\cS$
    \end{algorithmic}
\end{algorithm}

\section{Experiment}\label{sec:exp}
This section assesses the efficacy of our proposed method over SOTA methods across diverse datasets and neural architectures, followed by extensive ablation studies.

\begin{table*}[t]
    \centering
    \scalebox{0.79}{
        \begin{tabular}{@{}cc|ccccc|cccc@{}}
            \toprule[2pt]
                          &     & \multicolumn{5}{c|}{ConvNet} & \multicolumn{2}{c}{ResNet-18} & \multicolumn{2}{c}{ResNet-101}                                                                                                                                                    \\ \cmidrule(lr){3-7} \cmidrule(lr){8-9} \cmidrule(lr){10-11}
            Dataset       & IPC & MTT                          & IDM                           & TESLA                          & DATM                    & \algopt (Ours)          & SRe$^2$L       & \algopt (Ours)          & SRe$^2$L                & \algopt (Ours)          \\  \midrule
                          & 1   & 46.3 $\pm$ 0.8               & 45.6 $\pm$ 0.7                & \textbf{48.5 $\pm$ 0.8}        & 46.9 $\pm$ 0.5          & 23.5 $\pm$ 0.3          & 16.6 $\pm$ 0.9 & \textbf{22.9 $\pm$ 0.4} & 13.7 $\pm$ 0.2          & \textbf{18.7 $\pm$ 0.1} \\
            CIFAR10       & 10  & 65.3 $\pm$ 0.7               & 58.6 $\pm$ 0.1                & 66.4 $\pm$ 0.8                 & \textbf{66.8 $\pm$ 0.2} & 50.2 $\pm$ 0.3          & 29.3 $\pm$ 0.5 & \textbf{37.1 $\pm$ 0.3} & 24.3 $\pm$ 0.6          & \textbf{33.7 $\pm$ 0.3} \\
                          & 50  & 71.6 $\pm$ 0.2               & 67.5 $\pm$ 0.1                & 72.6 $\pm$ 0.7                 & \textbf{76.1 $\pm$ 0.3} & 68.4 $\pm$ 0.1          & 45.0 $\pm$ 0.7 & \textbf{62.1 $\pm$ 0.1} & 34.9 $\pm$ 0.1          & \textbf{51.6 $\pm$ 0.4} \\ \midrule
                          & 1   & 24.3 $\pm$ 0.3               & 20.1 $\pm$ 0.3                & 24.8 $\pm$ 0.5                 & \textbf{27.9 $\pm$ 0.2} & 19.6 $\pm$ 0.3          & 6.6 $\pm$ 0.2  & \textbf{11.0 $\pm$ 0.3} & 6.2 $\pm$ 0.0           & \textbf{10.8 $\pm$ 0.1} \\
            CIFAR-100     & 10  & 40.1 $\pm$ 0.4               & 45.1 $\pm$ 0.1                & 41.7 $\pm$ 0.3                 & 47.2 $\pm$ 0.4          & \textbf{48.1 $\pm$ 0.3} & 27.0 $\pm$ 0.4 & \textbf{42.6 $\pm$ 0.2} & 30.7 $\pm$ 0.3          & \textbf{41.1 $\pm$ 0.2} \\
                          & 50  & 47.7 $\pm$ 0.2               & 50.0 $\pm$ 0.2                & 47.9 $\pm$ 0.3                 & 55.0 $\pm$ 0.2          & \textbf{57.0 $\pm$ 0.1} & 50.2 $\pm$ 0.4 & \textbf{62.6 $\pm$ 0.1} & 56.9 $\pm$ 0.1          & \textbf{63.4 $\pm$ 0.3} \\ \midrule
                          & 1   & \textbf{47.7 $\pm$ 0.9}      & -                             & -                              & -                       & 33.8 $\pm$ 0.8          & 19.1 $\pm$ 1.1 & \textbf{35.8 $\pm$ 1.0} & 15.8 $\pm$ 0.6          & \textbf{25.1 $\pm$ 2.7} \\
            ImageNette    & 10  & 63.0 $\pm$ 1.3               & -                             & -                              & -                       & \textbf{63.2 $\pm$ 0.7} & 29.4 $\pm$ 3.0 & \textbf{61.4 $\pm$ 0.4} & 23.4 $\pm$ 0.8          & \textbf{54.0 $\pm$ 0.4} \\
                          & 50  & -                            & -                             & -                              & -                       & \textbf{83.8 $\pm$ 0.2} & 40.9 $\pm$ 0.3 & \textbf{80.4 $\pm$ 0.4} & 36.5 $\pm$ 0.7          & \textbf{75.0 $\pm$ 1.2} \\ \midrule
                          & 1   & \textbf{28.6 $\pm$ 0.8}      & -                             & -                              & -                       & 18.5 $\pm$ 0.9          & 13.3 $\pm$ 0.5 & \textbf{20.8 $\pm$ 1.2} & 13.4 $\pm$ 0.1          & \textbf{19.6 $\pm$ 1.8} \\
            ImageWoof     & 10  & 35.8 $\pm$ 1.8               & -                             & -                              & -                       & \textbf{40.6 $\pm$ 2.0} & 20.2 $\pm$ 0.2 & \textbf{38.5 $\pm$ 2.1} & 17.7 $\pm$ 0.9          & \textbf{31.3 $\pm$ 1.3} \\
                          & 50  & -                            & -                             & -                              & -                       & \textbf{61.5 $\pm$ 0.3} & 23.3 $\pm$ 0.3 & \textbf{68.5 $\pm$ 0.7} & 21.2 $\pm$ 0.2          & \textbf{59.1 $\pm$ 0.7} \\ \midrule
                          & 1   & 8.8 $\pm$ 0.3                & 10.1 $\pm$ 0.2                & -                              & \textbf{17.1 $\pm$ 0.3} & 12.0 $\pm$ 0.1          & 2.62 $\pm$ 0.1 & \textbf{9.7 $\pm$ 0.4}  & 1.9 $\pm$ 0.1           & \textbf{3.8 $\pm$ 0.1}  \\
            Tiny-ImageNet & 10  & 23.2 $\pm$ 0.2               & 21.9 $\pm$ 0.3                & -                              & 31.1 $\pm$ 0.3          & \textbf{39.6 $\pm$ 0.1} & 16.1 $\pm$ 0.2 & \textbf{41.9 $\pm$ 0.2} & 14.6 $\pm$ 1.1          & \textbf{22.9 $\pm$ 3.3} \\
                          & 50  & 28.0 $\pm$ 0.3               & 27.7 $\pm$ 0.3                & -                              & 39.7 $\pm$ 0.3          & \textbf{47.6 $\pm$ 0.2} & 41.1 $\pm$ 0.4 & \textbf{58.2 $\pm$ 0.1} & \textbf{42.5 $\pm$ 0.2} & 41.2 $\pm$ 0.4          \\ \midrule
                          & 1   & -                            & \textbf{11.2 $\pm$ 0.5}       & -                              & -                       & 7.1 $\pm$ 0.2           & 3.0 $\pm$ 0.3  & \textbf{8.1 $\pm$ 0.3}  & 2.1 $\pm$ 0.1           & \textbf{6.1 $\pm$ 0.8}  \\
            ImageNet-100  & 10  & -                            & 17.1 $\pm$ 0.6                & -                              & -                       & \textbf{29.6 $\pm$ 0.1} & 9.5 $\pm$ 0.4  & \textbf{36.0 $\pm$ 0.3} & 6.4 $\pm$ 0.1           & \textbf{33.9 $\pm$ 0.1} \\
                          & 50  & -                            & 26.3 $\pm$ 0.4                & -                              & -                       & \textbf{50.2 $\pm$ 0.2} & 27.0 $\pm$ 0.4 & \textbf{61.6 $\pm$ 0.1} & 25.7 $\pm$ 0.3          & \textbf{66.0 $\pm$ 0.6} \\ \midrule
                          & 1   & -                            & -                             & \textbf{7.7 $\pm$ 0.2}         & -                       & 6.4 $\pm$ 0.1           & 0.1 $\pm$ 0.1  & \textbf{6.6 $\pm$ 0.2}  & 0.6 $\pm$ 0.1           & \textbf{5.9 $\pm$ 0.4}  \\
            ImageNet-1K   & 10  & -                            & -                             & 17.8 $\pm$ 1.3                 & -                       & \textbf{20.4 $\pm$ 0.1} & 21.3 $\pm$ 0.6 & \textbf{42.0 $\pm$ 0.1} & 30.9 $\pm$ 0.1          & \textbf{48.3 $\pm$ 1.0} \\
                          & 50  & -                            & -                             & 27.9 $\pm$ 1.2                 & -                       & \textbf{38.4 $\pm$ 0.2} & 46.8 $\pm$ 0.2 & \textbf{56.5 $\pm$ 0.1} & 60.8 $\pm$ 0.5          & \textbf{61.2 $\pm$ 0.4} \\ \bottomrule[2pt]
        \end{tabular}
    }
    \vspace{-0.6em}
    \caption{
        \small
        \textbf{Comparison with the SOTA baseline dataset distillation methods.}
        We use identical neural networks for both dataset distillation and data evaluation.
        In general, following \cite{cazenavette2022dataset,cui2023scaling,zhao2023improved}, the ConvNet used for distillation are Conv-3 on CIFAR10 and CIFAR100, Conv-4 on Tiny-ImageNet and ImageNet-1K, Conv-5 on ImageNette and ImageWoof, Conv-6 on ImageNet-100.
        MTT and TESLA use a down-sampled version of image when distilling $224 \times 224$ images \cite{cazenavette2022dataset, cui2023scaling}.
        Following \cite{yin2023squeeze}, SRe$^2$L and \algopt use ResNet-18 to retrieve the distilled data, and evaluate on ResNet-18 and ResNet-101.
        Entries with ``-'' are absent due to scalability problems. See Appendix~\ref{sec:imple_details} for more details.
    }
    \vspace{-1.5em}
    \label{tb:main}
\end{table*}

\subsection{Experimental Setting} \label{sec:expset}
We list the settings below (see more details in Appendix~\ref{sec:aux_exp}).

\paragraph{Datasets.}
For low-resolution data ($32 \times 32$), we evaluate our method on two datasets, i.e., CIFAR-10 \cite{krizhevsky2009cifar} and CIFAR-100 \cite{krizhevsky2009learning}.
For high-resolution data, we conduct experiments on two large-scale datasets including Tiny-ImageNet ($64 \times 64$) \cite{le2015tiny} and full ImageNet-1K ($224 \times 224$) \cite{deng2009imagenet}.
Moreover, given the fact that most existing dataset distillation methods cannot be extended to large-scale high-resolution datasets, we further consider four widely used ImageNet-1K subsets in our evaluation: ImageNet-100~\cite{kim2022dataset}, ImageNette and ImageWoof~\cite{cazenavette2022dataset}.

\paragraph{Network architectures.}
Similar to the prior dataset distillation works \cite{yin2023squeeze,cazenavette2022dataset,zhao2023improved,cui2023scaling,guo2023towards}, we use ConvNet \cite{guo2023towards}, ResNet-18/ResNet-101 \cite{he2016deep}, EfficientNet-B0 \cite{tan2019efficientnet}, MobileNet-V2 \cite{sandler2018mobilenetv2}, as our backbone.

\paragraph{Baselines.}
We consider SOTA optimization-based dataset distillation methods that can scale to large high-resolution datasets for a broader practical impact:
\begin{itemize}
    \item MTT \cite{cazenavette2022dataset} is the first work that proposes trajectory matching-based dataset distillation, which can work on \emph{both low and high-resolution datasets}.
    \item IDM \cite{zhao2023improved} introduces an efficient dataset condensation method based on distribution matching, in contrast to computationally intensive optimization-oriented approaches \cite{zhao2020dataset,cazenavette2022dataset}, thus \emph{scaling to ImageNet-100}.
    \item TESLA \cite{cui2023scaling} is the first dataset distillation method \emph{scales to full ImageNet-1K}, which handles huge memory consumption of the MTT-based method with constant memory.
    \item DATM \cite{guo2023towards} is the first to \emph{outperform the original full dataset training performance} with large $\IPC$.
    \item SRe$^2$L \cite{yin2023squeeze} is a recent work to \emph{efficiently scale to ImageNet-1K}, and significantly outperforms existing methods on large high-resolution datasets.
          We consider it as our closest baseline.
\end{itemize}

\paragraph{Evaluation.}
Following previous research, we set \IPC to 1, 10, and 50.
To evaluate cross-architecture generalization, we use the distilled datasets from one neural architecture to train the other neural architectures from scratch and record the validation accuracy (see Table \ref{tb:crossarch}).
Furthermore, we evaluate the distillation efficiency in Table~\ref{tb:efficiency} by estimating the run-time cost of distilling the image, as well as the peak GPU memory usage.

\paragraph{Implementation details of \algopt.}
We employ a generalized configuration for $\cT^\prime$ (c.f. Section \ref{sec:infoextract} for definition), where the size $|\cT^\prime|$ is set as $300$.
We set $N=4$ (c.f. Section \ref{sec:inforecon} for definition) for high-resolution datasets and set $N=1$ for datasets with resolution less than $64 \times 64$.

\subsection{Main Results}

\paragraph{High-resolution datasets.}
To explore the potential of our approach for real-world applications, we first conduct experiments to compare with the state-of-the-art dataset distillation methods on Tiny-ImageNet and ImageNet-1K (including some subsets, e.g., ImageNet-100).
Table \ref{tb:main} demonstrates that \emph{our proposed method significantly outperforms existing methods or exhibits comparable results with large $\IPC = 10$ and $50$.}
However, when \IPC comes to 1, our approach struggles to effectively retain the information present in the original dataset, consequently leading to suboptimal outcomes.
\looseness=-1

\paragraph{Low-resolution datasets.}
To validate the robustness of our method across different-resolution datasets, we conduct more experiments on diminutive datasets such as CIFAR-10 and CIFAR-100 (see Table \ref{tb:main}).
Our \algopt demonstrates superior performance compared to conventional methods, particularly in scenarios involving larger distilled datasets such as CIFAR-100 with $\IPC = 50$.
However, similar to high-resolution scenarios, its efficacy diminishes when confronted with smaller datasets.

\begin{table}[t]
    \centering
    \scalebox{0.87}{
        \begin{tabular}{@{}cc|cc@{}}
            \toprule[2pt]
            \multicolumn{2}{c|}{Architecture} & Time Cost (ms) & Peak Memory (GB)                 \\ \midrule
            \multirow{2}{*}{ResNet-18}        & SRe$^2$L       & 2113.23          & 9.14          \\
                                              & Ours           & \textbf{39.89}   & \textbf{1.57} \\ \midrule
            \multirow{2}{*}{MobileNet-V2}     & SRe$^2$L       & 3783.16          & 12.93         \\
                                              & Ours           & \textbf{64.97}   & \textbf{2.35} \\ \midrule
            \multirow{2}{*}{EfficientNet-B0}  & SRe$^2$L       & 4412.42          & 11.92         \\
                                              & Ours           & \textbf{73.16}   & \textbf{2.34} \\ \bottomrule[2pt]
        \end{tabular}
    }
    \vspace{-0.6em}
    \caption{
        \small
        \textbf{Synthesis time and memory consumption ImageNet-1K.}
        We use a single RTX-4090 GPU for all methods to conduct experiments on ImageNet-1K.
        Time Cost represents the consumption (ms) for each image when generating 100 images simultaneously.
        Following the official implementation of SRe$^2$L \cite{yin2023squeeze}, the peak value of GPU memory usage is measured with a batch size of 100.
    }
    \vspace{-1.5em}
    \label{tb:efficiency}
\end{table}

\subsection{Efficiency Comparison}
Table \ref{tb:efficiency} distinctly showcases the \emph{superior efficiency of our dataset distillation approach in comparison to previous methodologies, demonstrating a significant performance advantage over state-of-the-art methods}.
Notably, we present a flexible peak memory scope, allowing dynamic adjustments to the batch size without compromising performance.
This efficiency is attributed to the fact that the primary memory consumption in our distillation procedure occurs exclusively during the scoring process of patches, while this process can be executed in parallel for images within a mini-batch\footnote{
    Conventional optimization-based dataset distillation methods \citep{yin2023squeeze,cazenavette2022dataset,guo2023towards} have to synthesize a batch of images simultaneously to guarantee its overall quality.
}.
Furthermore, the optimization-free nature of our \algopt ensures that the distillation time for an image is solely dependent on the scoring cost determined by the pre-trained teacher model size.
\looseness=-1

\subsection{Cross-architecture Generalization} \label{sec:crossarch}
To ensure the generalization capability of our distilled datasets, it is imperative to validate their effectiveness across multiple neural architectures not encountered when distilling datasets.
Table \ref{tb:crossarch} examines our \algopt with the SOTA SRe$^2$L \emph{and underscores the robust generalization ability of our method}.
Our success stems from two key aspects:
\begin{itemize}
    \item it enables high-realism distilled images (evidenced in \cite{cazenavette2023generalizing}).
    \item it exhibits insensitivity to variations in the teacher model.
\end{itemize}

\begin{table}[t]
    \setlength\tabcolsep{1.5pt}
    \centering
    \scalebox{0.86}{
        \begin{tabular}{@{}cc|ccc@{}}
            \toprule[2pt]
            \multicolumn{2}{c|}{Verifier\textbackslash{}Observer} & ResNet-18 & EfficientNet-B0         & MobileNet-V2                                      \\ \midrule
            \multirow{2}{*}{ResNet-18}                            & SRe$^2$L  & 21.7 $\pm$ 0.6          & 11.7 $\pm$ 0.2          & 15.4 $\pm$ 0.2          \\
                                                                  & Ours      & \textbf{42.3 $\pm$ 0.6} & \textbf{31.0 $\pm$ 0.1} & \textbf{40.4 $\pm$ 0.1} \\ \midrule
            \multirow{2}{*}{MobileNet-V2}                         & SRe$^2$L  & 19.7 $\pm$ 0.1          & 9.8 $\pm$ 0.4           & 10.2 $\pm$ 2.6          \\
                                                                  & Ours      & \textbf{34.4 $\pm$ 0.2} & \textbf{24.1 $\pm$ 0.8} & \textbf{33.8 $\pm$ 0.6} \\ \midrule
            \multirow{2}{*}{EfficientNet-B0}                      & SRe$^2$L  & 25.2 $\pm$ 0.2          & 11.4 $\pm$ 2.5          & 20.5 $\pm$ 0.2          \\
                                                                  & Ours      & \textbf{42.8 $\pm$ 0.5} & \textbf{33.3 $\pm$ 0.9} & \textbf{43.6 $\pm$ 0.2} \\ \bottomrule[2pt]
        \end{tabular}
    }
    \vspace{-0.6em}
    \caption{
        \small
        \textbf{Evaluating ImageNet-1K top-1 accuracy on cross-architecture generalization.}
        Distill dataset with ResNet-18, EfficientNet-B0, and MobileNet-V2, and then versus transfer to each other architecture.
        We can not conduct experiments for SRe$^2$L when the model using for distillation without batch normalization, which necessitates \cite{yin2023squeeze}.
        All methods are evaluated with $\IPC=10$.
    }
    \vspace{-0.5em}
    \label{tb:crossarch}
\end{table}

\subsection{Ablation Study} \label{sec:ablationstudy}
The effectiveness of \algopt hinges on two pivotal factors: the size $|\cT_c^\prime|$ of pre-selected subset $\cT_c^\prime$ (c.f.\ Section \ref{sec:infoextract}) and the number of patches $N$ per distilled image (defined in Section \ref{sec:inforecon}).
In this section, we set $\IPC = 10$ and employ ResNet-18 as the network backbone to examine how these factors influence the diversity and realism of the distilled dataset (see Appendix~\ref{sec:aux_ablationstudy} for investigation on more factors).

\begin{figure}[t]
    \begin{minipage}{0.20\textwidth}
        \centering
        \includegraphics[width=1.25\linewidth]{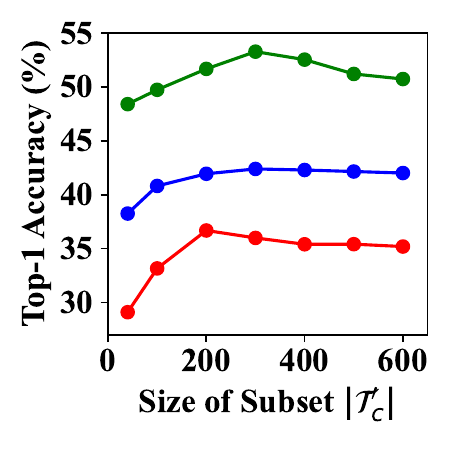}
    \end{minipage}
    \hspace{0.6cm}
    \begin{minipage}{0.20\textwidth}
        \centering
        \includegraphics[width=1.15\linewidth]{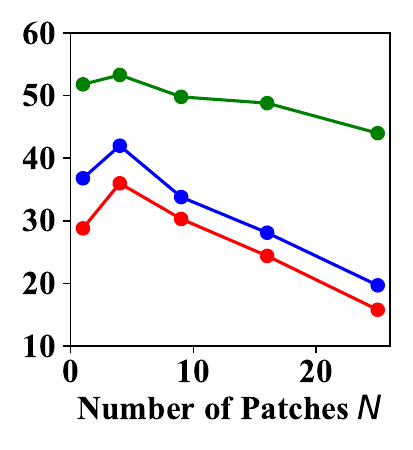}
    \end{minipage}
    \vspace{-1.5em}
    \caption{
        \textbf{Ablation study on $\abs{\cT_c^\prime}$ and $N$}, i.e., the pre-selected subset size $\cT_c^\prime$ (left), and the number of patches $N$ within each distilled image (right).
        The emerald \textcolor{emerald}{$\bullet$}, red \textcolor{red}{$\bullet$}, and blue \textcolor{blue}{$\bullet$} denote ImageNet-10, ImageNet-100, and ImageNet-1K respectively.
        \looseness=-1
    }
    \vspace{-1.75em}
    \label{fig:elements}
\end{figure}

\paragraph{On the impact of pre-selected subset size $|\cT_c^\prime|$.}
The experimental results in Figure \ref{fig:elements} gives a more intuitive demonstration on the impact of $|\cT_c^\prime|$, alongside the discussion in Section \ref{sec:infoextract}:
\begin{itemize}
    \item The performance abruptly drops when $|\cT_c^\prime|$ is equal to $N \times \IPC$, i.e., the Stage 1 in our Algorithm \ref{alg:framework} becomes the simple uniform random sampling.
          In this case, the diversity is maximized but the realism is poor, thus resulting in catastrophically degraded performance.
    \item As $|\cT_c^\prime|$ continuously increases and exceeds a threshold, our framework collects more realistic images from $\cT_c^\prime$ but their patterns may be repeated, thus hurting diversity and consequent performance.
\end{itemize}
Therefore, a proper $|\cT_c^\prime|$ would balance the trade-off between data diversity and realism.
In this instance, a value approaching $300$ maximizes the total sum for our target in Equation \eqref{eq:vinfo}, as indicated in Figure \ref{fig:elements}.

\paragraph{On the impact of squeezing $N$ patches into one distilled image.}
The number of patches $N$ shares similar patterns as that of $|\cT_c^\prime|$.
Specifically, though we can compress more patches from $\cT$ into a distilled dataset $\cS$ by increasing $N$ increases to benefit the data diversity, it also results in a lower resolution for the source patches (see our explanation in Footnote \ref{fn:patchesnumber}), thus hurting the realism.
Therefore, a proper number of patches $N$ is important to achieve our objective in \eqref{eq:principle}.
Figure \ref{fig:elements} showcases that the validation performance rises to the highest on selected three datasets when $N=4$.

\section{Conclusion}
In this work, we introduce an optimization-free and efficient paradigm which successfully distills a dataset with $\IPC = 10$ from the entirety of ImageNet-1K, concurrently achieving $42\%$ top-1 validation accuracy with ResNet-18.
Furthermore, our method exhibits robust cross-architecture generalization, surpassing SOTA method by a factor of $2\times$ in performance.

\section*{Acknowledgement}
We thank Xinyi Shang, Zexi Li and anonymous reviewers for their precious comments and feedback.
This work was supported in part by the National Science and Technology Major Project (No.\ 2022ZD0115101), the Research Center for Industries of the Future (RCIF) at Westlake University, and the Westlake Education Foundation.

  {
    \small
    \bibliographystyle{configuration/ieeenat_fullname}
    \bibliography{paper}

\begin{thebibliography}{53}
\providecommand{\natexlab}[1]{#1}
\providecommand{\url}[1]{\texttt{#1}}
\expandafter\ifx\csname urlstyle\endcsname\relax
  \providecommand{\doi}[1]{doi: #1}\else
  \providecommand{\doi}{doi: \begingroup \urlstyle{rm}\Url}\fi

\bibitem[Cazenavette et~al.(2022)Cazenavette, Wang, Torralba, Efros, and Zhu]{cazenavette2022dataset}
George Cazenavette, Tongzhou Wang, Antonio Torralba, Alexei~A Efros, and Jun-Yan Zhu.
\newblock Dataset distillation by matching training trajectories.
\newblock In \emph{Proceedings of the IEEE/CVF Conference on Computer Vision and Pattern Recognition}, pages 4750--4759, 2022.

\bibitem[Cazenavette et~al.(2023)Cazenavette, Wang, Torralba, Efros, and Zhu]{cazenavette2023generalizing}
George Cazenavette, Tongzhou Wang, Antonio Torralba, Alexei~A Efros, and Jun-Yan Zhu.
\newblock Generalizing dataset distillation via deep generative prior.
\newblock In \emph{Proceedings of the IEEE/CVF Conference on Computer Vision and Pattern Recognition}, pages 3739--3748, 2023.

\bibitem[Coleman et~al.(2019)Coleman, Yeh, Mussmann, Mirzasoleiman, Bailis, Liang, Leskovec, and Zaharia]{coleman2019selection}
Cody Coleman, Christopher Yeh, Stephen Mussmann, Baharan Mirzasoleiman, Peter Bailis, Percy Liang, Jure Leskovec, and Matei Zaharia.
\newblock Selection via proxy: Efficient data selection for deep learning.
\newblock \emph{arXiv preprint arXiv:1906.11829}, 2019.

\bibitem[Cui et~al.(2022)Cui, Wang, Si, and Hsieh]{cui2022dc}
Justin Cui, Ruochen Wang, Si Si, and Cho-Jui Hsieh.
\newblock Dc-bench: Dataset condensation benchmark.
\newblock \emph{Advances in Neural Information Processing Systems}, 35:\penalty0 810--822, 2022.

\bibitem[Cui et~al.(2023)Cui, Wang, Si, and Hsieh]{cui2023scaling}
Justin Cui, Ruochen Wang, Si Si, and Cho-Jui Hsieh.
\newblock Scaling up dataset distillation to imagenet-1k with constant memory.
\newblock In \emph{International Conference on Machine Learning}, pages 6565--6590. PMLR, 2023.

\bibitem[Deng et~al.(2009)Deng, Dong, Socher, Li, Li, and Fei-Fei]{deng2009imagenet}
Jia Deng, Wei Dong, Richard Socher, Li-Jia Li, Kai Li, and Li Fei-Fei.
\newblock Imagenet: A large-scale hierarchical image database.
\newblock In \emph{2009 IEEE conference on computer vision and pattern recognition}, pages 248--255. Ieee, 2009.

\bibitem[Dosovitskiy et~al.(2021)Dosovitskiy, Beyer, Kolesnikov, Weissenborn, Zhai, Unterthiner, Dehghani, Minderer, Heigold, Gelly, Uszkoreit, and Houlsby]{dosovitskiy2021an}
Alexey Dosovitskiy, Lucas Beyer, Alexander Kolesnikov, Dirk Weissenborn, Xiaohua Zhai, Thomas Unterthiner, Mostafa Dehghani, Matthias Minderer, Georg Heigold, Sylvain Gelly, Jakob Uszkoreit, and Neil Houlsby.
\newblock An image is worth 16x16 words: Transformers for image recognition at scale.
\newblock In \emph{International Conference on Learning Representations}, 2021.

\bibitem[Du et~al.(2023)Du, Jiang, Tan, Zhou, and Li]{du2023minimizing}
Jiawei Du, Yidi Jiang, Vincent~YF Tan, Joey~Tianyi Zhou, and Haizhou Li.
\newblock Minimizing the accumulated trajectory error to improve dataset distillation.
\newblock In \emph{Proceedings of the IEEE/CVF Conference on Computer Vision and Pattern Recognition}, pages 3749--3758, 2023.

\bibitem[Ethayarajh et~al.(2022)Ethayarajh, Choi, and Swayamdipta]{ethayarajh2022understanding}
Kawin Ethayarajh, Yejin Choi, and Swabha Swayamdipta.
\newblock Understanding dataset difficulty with $\mathcal{V}$-usable information.
\newblock In \emph{International Conference on Machine Learning}, pages 5988--6008. PMLR, 2022.

\bibitem[Feldman and Zhang(2020)]{feldman2020neural}
Vitaly Feldman and Chiyuan Zhang.
\newblock What neural networks memorize and why: Discovering the long tail via influence estimation.
\newblock \emph{Advances in Neural Information Processing Systems}, 33:\penalty0 2881--2891, 2020.

\bibitem[Forgy(1965)]{forgy1965cluster}
Edward~W Forgy.
\newblock Cluster analysis of multivariate data: efficiency versus interpretability of classifications.
\newblock \emph{biometrics}, 21:\penalty0 768--769, 1965.

\bibitem[Goodfellow et~al.(2014)Goodfellow, Pouget-Abadie, Mirza, Xu, Warde-Farley, Ozair, Courville, and Bengio]{goodfellow2014generative}
Ian Goodfellow, Jean Pouget-Abadie, Mehdi Mirza, Bing Xu, David Warde-Farley, Sherjil Ozair, Aaron Courville, and Yoshua Bengio.
\newblock Generative adversarial nets.
\newblock \emph{Advances in neural information processing systems}, 27, 2014.

\bibitem[Guo et~al.(2023)Guo, Wang, Cazenavette, Li, Zhang, and You]{guo2023towards}
Ziyao Guo, Kai Wang, George Cazenavette, Hui Li, Kaipeng Zhang, and Yang You.
\newblock Towards lossless dataset distillation via difficulty-aligned trajectory matching.
\newblock \emph{arXiv preprint arXiv:2310.05773}, 2023.

\bibitem[He et~al.(2016)He, Zhang, Ren, and Sun]{he2016deep}
Kaiming He, Xiangyu Zhang, Shaoqing Ren, and Jian Sun.
\newblock Deep residual learning for image recognition.
\newblock In \emph{Proceedings of the IEEE conference on computer vision and pattern recognition}, pages 770--778, 2016.

\bibitem[He et~al.(2024)He, Xiao, Zhou, and Tsang]{he2024multisize}
Yang He, Lingao Xiao, Joey~Tianyi Zhou, and Ivor Tsang.
\newblock Multisize dataset condensation.
\newblock \emph{arXiv preprint arXiv:2403.06075}, 2024.

\bibitem[Ioffe and Szegedy(2015)]{ioffe2015batch}
Sergey Ioffe and Christian Szegedy.
\newblock Batch normalization: Accelerating deep network training by reducing internal covariate shift.
\newblock In \emph{International conference on machine learning}, pages 448--456. pmlr, 2015.

\bibitem[Kaplan et~al.(2020)Kaplan, McCandlish, Henighan, Brown, Chess, Child, Gray, Radford, Wu, and Amodei]{kaplan2020scaling}
Jared Kaplan, Sam McCandlish, Tom Henighan, Tom~B Brown, Benjamin Chess, Rewon Child, Scott Gray, Alec Radford, Jeffrey Wu, and Dario Amodei.
\newblock Scaling laws for neural language models.
\newblock \emph{arXiv preprint arXiv:2001.08361}, 2020.

\bibitem[Kim et~al.(2022)Kim, Kim, Oh, Yun, Song, Jeong, Ha, and Song]{kim2022dataset}
Jang-Hyun Kim, Jinuk Kim, Seong~Joon Oh, Sangdoo Yun, Hwanjun Song, Joonhyun Jeong, Jung-Woo Ha, and Hyun~Oh Song.
\newblock Dataset condensation via efficient synthetic-data parameterization.
\newblock In \emph{International Conference on Machine Learning}, pages 11102--11118. PMLR, 2022.

\bibitem[Krizhevsky et~al.(2009{\natexlab{a}})Krizhevsky, Hinton, et~al.]{krizhevsky2009learning}
Alex Krizhevsky, Geoffrey Hinton, et~al.
\newblock Learning multiple layers of features from tiny images.
\newblock 2009{\natexlab{a}}.

\bibitem[Krizhevsky et~al.(2009{\natexlab{b}})Krizhevsky, Nair, and Hinton]{krizhevsky2009cifar}
Alex Krizhevsky, Vinod Nair, and Geoffrey Hinton.
\newblock Cifar-10 and cifar-100 datasets.
\newblock \emph{URl: https://www. cs. toronto. edu/kriz/cifar. html}, 6\penalty0 (1):\penalty0 1, 2009{\natexlab{b}}.

\bibitem[Le and Yang(2015)]{le2015tiny}
Ya Le and Xuan Yang.
\newblock Tiny imagenet visual recognition challenge.
\newblock \emph{CS 231N}, 7\penalty0 (7):\penalty0 3, 2015.

\bibitem[Liu et~al.(2022)Liu, Hu, Lin, Yao, Xie, Wei, Ning, Cao, Zhang, Dong, et~al.]{liu2022swin}
Ze Liu, Han Hu, Yutong Lin, Zhuliang Yao, Zhenda Xie, Yixuan Wei, Jia Ning, Yue Cao, Zheng Zhang, Li Dong, et~al.
\newblock Swin transformer v2: Scaling up capacity and resolution.
\newblock In \emph{Proceedings of the IEEE/CVF conference on computer vision and pattern recognition}, pages 12009--12019, 2022.

\bibitem[Loo et~al.(2022)Loo, Hasani, Amini, and Rus]{loo2022efficient}
Noel Loo, Ramin Hasani, Alexander Amini, and Daniela Rus.
\newblock Efficient dataset distillation using random feature approximation.
\newblock \emph{Advances in Neural Information Processing Systems}, 35:\penalty0 13877--13891, 2022.

\bibitem[Ma et~al.(2022)Ma, Tsao, and Shum]{ma2022principles}
Yi Ma, Doris Tsao, and Heung-Yeung Shum.
\newblock On the principles of parsimony and self-consistency for the emergence of intelligence.
\newblock \emph{Frontiers of Information Technology \& Electronic Engineering}, 23\penalty0 (9):\penalty0 1298--1323, 2022.

\bibitem[Meding et~al.(2021)Meding, Buschoff, Geirhos, and Wichmann]{meding2021trivial}
Kristof Meding, Luca M~Schulze Buschoff, Robert Geirhos, and Felix~A Wichmann.
\newblock Trivial or impossible--dichotomous data difficulty masks model differences (on imagenet and beyond).
\newblock \emph{arXiv preprint arXiv:2110.05922}, 2021.

\bibitem[Oord et~al.(2018)Oord, Li, and Vinyals]{oord2018representation}
Aaron van~den Oord, Yazhe Li, and Oriol Vinyals.
\newblock Representation learning with contrastive predictive coding.
\newblock \emph{arXiv preprint arXiv:1807.03748}, 2018.

\bibitem[Paul et~al.(2021)Paul, Ganguli, and Dziugaite]{paul2021deep}
Mansheej Paul, Surya Ganguli, and Gintare~Karolina Dziugaite.
\newblock Deep learning on a data diet: Finding important examples early in training.
\newblock \emph{Advances in Neural Information Processing Systems}, 34:\penalty0 20596--20607, 2021.

\bibitem[Radford et~al.(2021)Radford, Kim, Hallacy, Ramesh, Goh, Agarwal, Sastry, Askell, Mishkin, Clark, et~al.]{radford2021learning}
Alec Radford, Jong~Wook Kim, Chris Hallacy, Aditya Ramesh, Gabriel Goh, Sandhini Agarwal, Girish Sastry, Amanda Askell, Pamela Mishkin, Jack Clark, et~al.
\newblock Learning transferable visual models from natural language supervision.
\newblock In \emph{International conference on machine learning}, pages 8748--8763. PMLR, 2021.

\bibitem[Sandler et~al.(2018)Sandler, Howard, Zhu, Zhmoginov, and Chen]{sandler2018mobilenetv2}
Mark Sandler, Andrew Howard, Menglong Zhu, Andrey Zhmoginov, and Liang-Chieh Chen.
\newblock Mobilenetv2: Inverted residuals and linear bottlenecks.
\newblock In \emph{Proceedings of the IEEE conference on computer vision and pattern recognition}, pages 4510--4520, 2018.

\bibitem[Shannon(1948)]{shannon1948mathematical}
Claude~Elwood Shannon.
\newblock A mathematical theory of communication.
\newblock \emph{The Bell system technical journal}, 27\penalty0 (3):\penalty0 379--423, 1948.

\bibitem[Shao et~al.(2023)Shao, Yin, Zhou, Zhang, and Shen]{shao2023generalized}
Shitong Shao, Zeyuan Yin, Muxin Zhou, Xindong Zhang, and Zhiqiang Shen.
\newblock Generalized large-scale data condensation via various backbone and statistical matching.
\newblock \emph{arXiv preprint arXiv:2311.17950}, 2023.

\bibitem[Shen and Xing(2022)]{shen2022fast}
Zhiqiang Shen and Eric Xing.
\newblock A fast knowledge distillation framework for visual recognition.
\newblock In \emph{European Conference on Computer Vision}, pages 673--690. Springer, 2022.

\bibitem[Simonyan and Zisserman(2014)]{simonyan2014very}
Karen Simonyan and Andrew Zisserman.
\newblock Very deep convolutional networks for large-scale image recognition.
\newblock \emph{arXiv preprint arXiv:1409.1556}, 2014.

\bibitem[Sorscher et~al.(2022)Sorscher, Geirhos, Shekhar, Ganguli, and Morcos]{sorscher2022beyond}
Ben Sorscher, Robert Geirhos, Shashank Shekhar, Surya Ganguli, and Ari Morcos.
\newblock Beyond neural scaling laws: beating power law scaling via data pruning.
\newblock \emph{Advances in Neural Information Processing Systems}, 35:\penalty0 19523--19536, 2022.

\bibitem[Tan et~al.(2023)Tan, Wu, Du, Chen, Wang, Wang, and Qi]{tan2023data}
Haoru Tan, Sitong Wu, Fei Du, Yukang Chen, Zhibin Wang, Fan Wang, and Xiaojuan Qi.
\newblock Data pruning via moving-one-sample-out.
\newblock \emph{arXiv preprint arXiv:2310.14664}, 2023.

\bibitem[Tan and Le(2019)]{tan2019efficientnet}
Mingxing Tan and Quoc Le.
\newblock Efficientnet: Rethinking model scaling for convolutional neural networks.
\newblock In \emph{International conference on machine learning}, pages 6105--6114. PMLR, 2019.

\bibitem[Toneva et~al.(2018)Toneva, Sordoni, Combes, Trischler, Bengio, and Gordon]{toneva2018empirical}
Mariya Toneva, Alessandro Sordoni, Remi Tachet~des Combes, Adam Trischler, Yoshua Bengio, and Geoffrey~J Gordon.
\newblock An empirical study of example forgetting during deep neural network learning.
\newblock \emph{arXiv preprint arXiv:1812.05159}, 2018.

\bibitem[Wang et~al.(2022)Wang, Zhao, Peng, Zhu, Yang, Wang, Huang, Bilen, Wang, and You]{wang2022cafe}
Kai Wang, Bo Zhao, Xiangyu Peng, Zheng Zhu, Shuo Yang, Shuo Wang, Guan Huang, Hakan Bilen, Xinchao Wang, and Yang You.
\newblock Cafe: Learning to condense dataset by aligning features.
\newblock In \emph{Proceedings of the IEEE/CVF Conference on Computer Vision and Pattern Recognition}, pages 12196--12205, 2022.

\bibitem[Wang et~al.(2018)Wang, Zhu, Torralba, and Efros]{wang2018dataset}
Tongzhou Wang, Jun-Yan Zhu, Antonio Torralba, and Alexei~A Efros.
\newblock Dataset distillation.
\newblock \emph{arXiv preprint arXiv:1811.10959}, 2018.

\bibitem[Welling(2009)]{welling2009herding}
Max Welling.
\newblock Herding dynamical weights to learn.
\newblock In \emph{Proceedings of the 26th Annual International Conference on Machine Learning}, pages 1121--1128, 2009.

\bibitem[Wong et~al.(2016)Wong, Gatt, Stamatescu, and McDonnell]{wong2016understanding}
Sebastien~C Wong, Adam Gatt, Victor Stamatescu, and Mark~D McDonnell.
\newblock Understanding data augmentation for classification: when to warp?
\newblock In \emph{2016 international conference on digital image computing: techniques and applications (DICTA)}, pages 1--6. IEEE, 2016.

\bibitem[Xu et~al.(2020)Xu, Zhao, Song, Stewart, and Ermon]{xu2020theory}
Yilun Xu, Shengjia Zhao, Jiaming Song, Russell Stewart, and Stefano Ermon.
\newblock A theory of usable information under computational constraints.
\newblock \emph{arXiv preprint arXiv:2002.10689}, 2020.

\bibitem[Yin et~al.(2020)Yin, Molchanov, Alvarez, Li, Mallya, Hoiem, Jha, and Kautz]{yin2020dreaming}
Hongxu Yin, Pavlo Molchanov, Jose~M Alvarez, Zhizhong Li, Arun Mallya, Derek Hoiem, Niraj~K Jha, and Jan Kautz.
\newblock Dreaming to distill: Data-free knowledge transfer via deepinversion.
\newblock In \emph{Proceedings of the IEEE/CVF Conference on Computer Vision and Pattern Recognition}, pages 8715--8724, 2020.

\bibitem[Yin et~al.(2023)Yin, Xing, and Shen]{yin2023squeeze}
Zeyuan Yin, Eric Xing, and Zhiqiang Shen.
\newblock Squeeze, recover and relabel: Dataset condensation at imagenet scale from a new perspective.
\newblock \emph{arXiv preprint arXiv:2306.13092}, 2023.

\bibitem[Yu et~al.(2023)Yu, Liu, and Wang]{yu2023dataset}
Ruonan Yu, Songhua Liu, and Xinchao Wang.
\newblock Dataset distillation: A comprehensive review.
\newblock \emph{arXiv preprint arXiv:2301.07014}, 2023.

\bibitem[Yun et~al.(2019)Yun, Han, Oh, Chun, Choe, and Yoo]{yun2019cutmix}
Sangdoo Yun, Dongyoon Han, Seong~Joon Oh, Sanghyuk Chun, Junsuk Choe, and Youngjoon Yoo.
\newblock Cutmix: Regularization strategy to train strong classifiers with localizable features.
\newblock In \emph{Proceedings of the IEEE/CVF international conference on computer vision}, pages 6023--6032, 2019.

\bibitem[Yun et~al.(2021)Yun, Oh, Heo, Han, Choe, and Chun]{yun2021re}
Sangdoo Yun, Seong~Joon Oh, Byeongho Heo, Dongyoon Han, Junsuk Choe, and Sanghyuk Chun.
\newblock Re-labeling imagenet: from single to multi-labels, from global to localized labels.
\newblock In \emph{Proceedings of the IEEE/CVF Conference on Computer Vision and Pattern Recognition}, pages 2340--2350, 2021.

\bibitem[Zagoruyko and Komodakis(2016)]{zagoruyko2016paying}
Sergey Zagoruyko and Nikos Komodakis.
\newblock Paying more attention to attention: Improving the performance of convolutional neural networks via attention transfer.
\newblock \emph{arXiv preprint arXiv:1612.03928}, 2016.

\bibitem[Zhang et~al.(2023)Zhang, Zhang, Lei, Mukherjee, Pan, Zhao, Ding, Li, and Xu]{zhang2023accelerating}
Lei Zhang, Jie Zhang, Bowen Lei, Subhabrata Mukherjee, Xiang Pan, Bo Zhao, Caiwen Ding, Yao Li, and Dongkuan Xu.
\newblock Accelerating dataset distillation via model augmentation.
\newblock In \emph{Proceedings of the IEEE/CVF Conference on Computer Vision and Pattern Recognition}, pages 11950--11959, 2023.

\bibitem[Zhao and Bilen(2023)]{zhao2023dataset}
Bo Zhao and Hakan Bilen.
\newblock Dataset condensation with distribution matching.
\newblock In \emph{Proceedings of the IEEE/CVF Winter Conference on Applications of Computer Vision}, pages 6514--6523, 2023.

\bibitem[Zhao et~al.(2020)Zhao, Mopuri, and Bilen]{zhao2020dataset}
Bo Zhao, Konda~Reddy Mopuri, and Hakan Bilen.
\newblock Dataset condensation with gradient matching.
\newblock \emph{arXiv preprint arXiv:2006.05929}, 2020.

\bibitem[Zhao et~al.(2023)Zhao, Li, Qin, and Yu]{zhao2023improved}
Ganlong Zhao, Guanbin Li, Yipeng Qin, and Yizhou Yu.
\newblock Improved distribution matching for dataset condensation.
\newblock In \emph{Proceedings of the IEEE/CVF Conference on Computer Vision and Pattern Recognition}, pages 7856--7865, 2023.

\bibitem[Zhou et~al.(2022)Zhou, Nezhadarya, and Ba]{zhou2022dataset}
Yongchao Zhou, Ehsan Nezhadarya, and Jimmy Ba.
\newblock Dataset distillation using neural feature regression.
\newblock \emph{Advances in Neural Information Processing Systems}, 35:\penalty0 9813--9827, 2022.

\end{thebibliography}
  }

\clearpage
\setcounter{page}{1}
\maketitlesupplementary

\appendix

\section{Distilled Images Comparison}
\label{sec:aux_visualization}
\begin{figure*}
    \centering
    \begin{subfigure}{0.98\linewidth}
        \includegraphics[width=17cm]{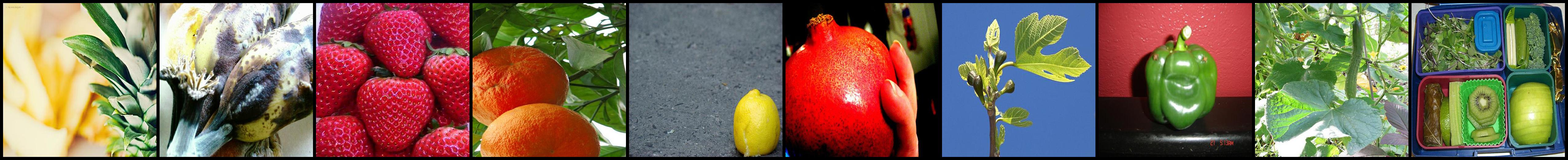}
        \caption{Random selection of original dataset}
    \end{subfigure}
    \vfill
    \begin{subfigure}{0.98\linewidth}
        \includegraphics[width=17cm]{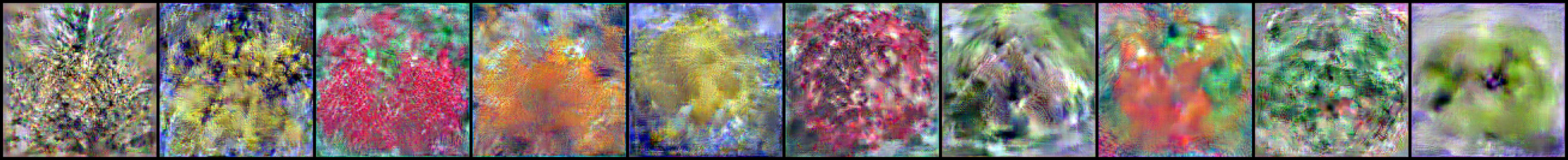}
        \caption{MTT~\cite{cazenavette2022dataset}}
    \end{subfigure}
    \vfill
    \begin{subfigure}{0.98\linewidth}
        \includegraphics[width=17cm]{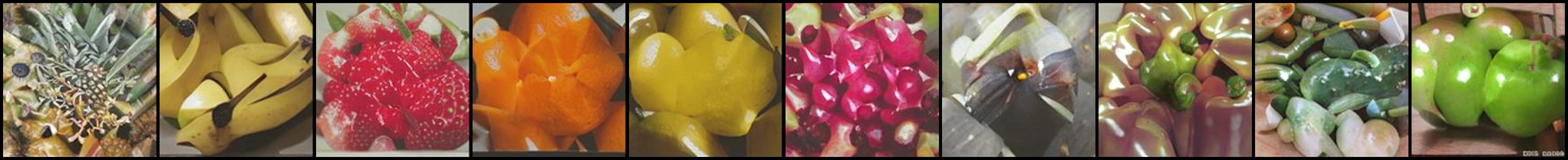}
        \caption{GLaD~\cite{cazenavette2023generalizing}}
    \end{subfigure}
    \vfill
    \begin{subfigure}{0.98\linewidth}
        \includegraphics[width=17cm]{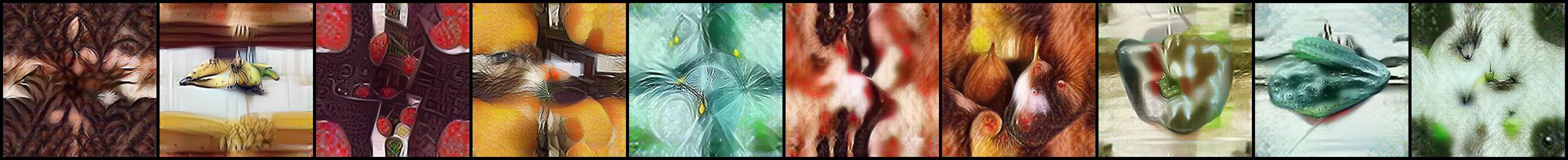}
        \caption{SRe$^2$L~\cite{yin2023squeeze}}
    \end{subfigure}
    \vfill
    \begin{subfigure}{0.98\linewidth}
        \includegraphics[width=17cm]{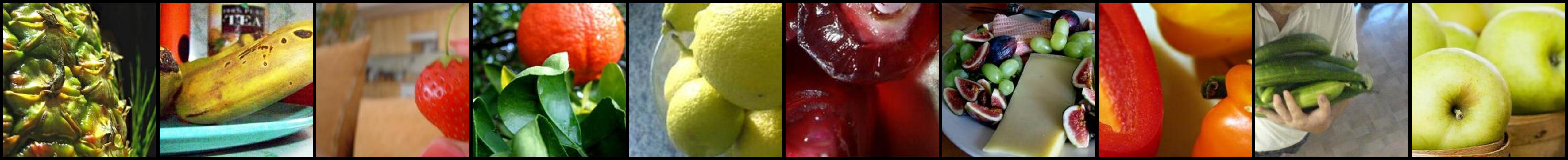}
        \caption{Herding~\cite{welling2009herding}}
    \end{subfigure}
    \vfill
    \begin{subfigure}{0.98\linewidth}
        \includegraphics[width=17cm]{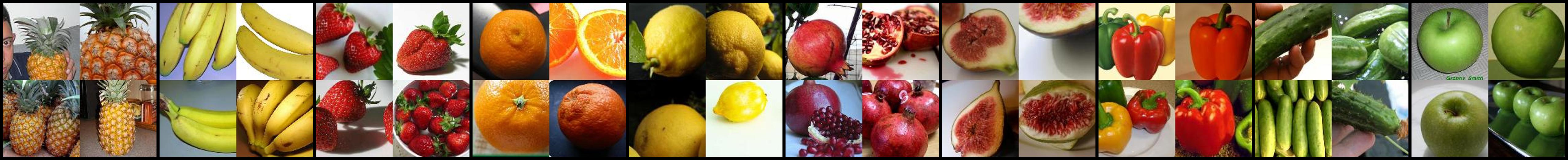}
        \caption{\algopt (Ours)}
    \end{subfigure}
    \caption{
        \small
        \textbf{Visualization of images synthesized using various dataset distillation methods}.
        We consider the ImageNet-Fruits \cite{cazenavette2022dataset} dataset, comprising a total of 10 distinct fruit types.
    }
    \vspace{-1.5em}
    \label{fig:comparison_more}
\end{figure*}

\begin{figure*}
    \centering
    \begin{subfigure}{0.98\linewidth}
        \includegraphics[width=17cm]{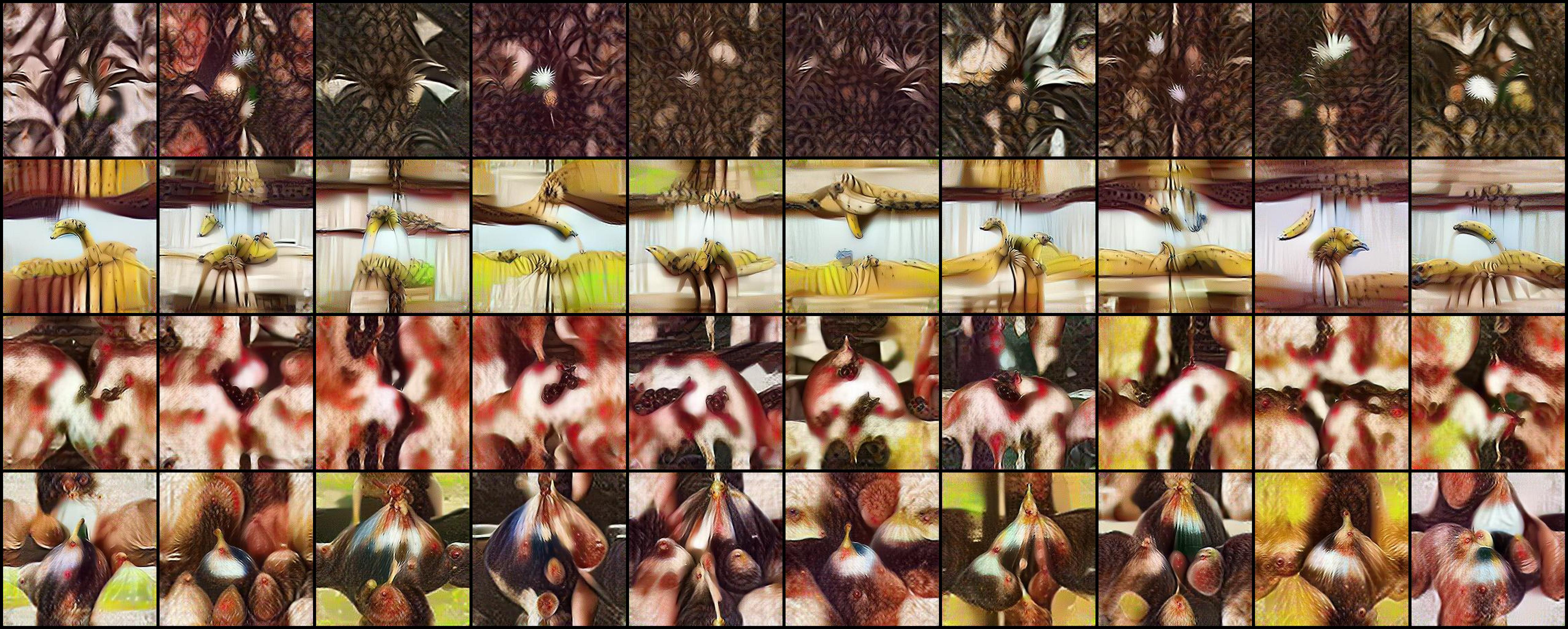}
        \caption{SRe$^2$L~\cite{yin2023squeeze}}
    \end{subfigure}
    \vfill
    \begin{subfigure}{0.98\linewidth}
        \includegraphics[width=17cm]{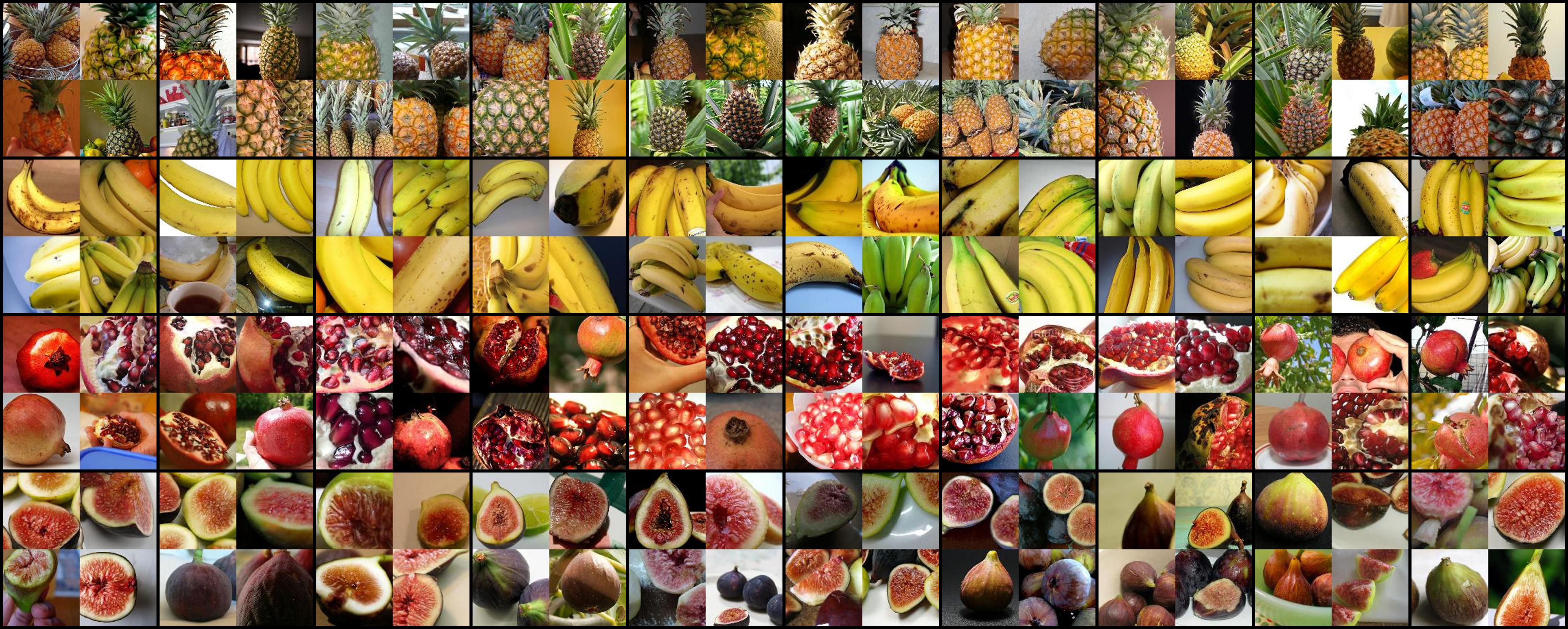}
        \caption{\algopt (Ours)}
    \end{subfigure}
    \caption{
        \small
        \textbf{Visualization of images synthesized using two dataset distillation methods}.
        We consider a subset of the ImageNet-Fruits \cite{cazenavette2022dataset} dataset, comprising a total of 4 distinct fruit types.
    }
    \vspace{-1.5em}
    \label{fig:comparison_sqe2l_rded}
\end{figure*}

Several additional examples of distilled images are presented in Figure~\ref{fig:comparison_more}.
Besides, we conduct a meticulous comparison between our proposed \algopt and the closest approach, SRe$^2$L.
The distilled images generated by SRe$^2$L are scrutinized in Figure~\ref{fig:comparison_sqe2l_rded}, revealing two noteworthy observations:
\begin{itemize}
    \item SRe$^2$L exhibits a limitation in generating diverse features within each distilled image.
    \item The diversity and realism of distilled images within each class are notably lacking.
\end{itemize}
In contrast, our proposed method, \algopt, demonstrates a superior capability to achieve high diversity in both the features within individual images and across images within each class, all while maintaining a high level of realism.

\section{$\cV$-information Theory}
\label{sec:aux_vinfo}

\subsection{Definitions}

The following definitions, as outlined by \citet{xu2020theory}, establish the groundwork for our discussion:

\begin{definition}[Predictive Family]
    \label{def:predfamily}
    Let $\Omega=\{f:\mathcal{X} \cup \lbrace \varnothing \rbrace \to \mathcal{P}(\mathcal{Y})\}$. $\cV \subseteq \Omega$ is a predictive family if it satisfies
    \begin{equation}
        \forall f \in \cV, \forall P \in \mathrm{range}(f), \quad \exists f' \in \cV,
    \end{equation} \label{eq:optional-ignorance}
    $s.t. \quad \forall x \in X, f'[x] = P, f'[\varnothing] = P$.
\end{definition}

A predictive family denotes a collection of permissible predictive models (observers) available to an agent, often constrained by computational or statistical limitations.
\citet{xu2020theory} term the supplementary criterion in (\ref{eq:optional-ignorance}) as \emph{optional ignorance}.
In essence, this implies that within the framework of the subsequent prediction game we delineate, the agent possesses the discretion to disregard the provided side information at thier discretion.

\begin{definition}
    Consider random variables $X$ and $Y$ with corresponding sample spaces $\mathcal{X}$ and $\mathcal{Y}$. Let $\varnothing$ denote a null input that imparts no information about $Y$.
    Within the context of a predictive family $\cV \subseteq \Omega = \{ f: \mathcal{X} \cup \varnothing \to \cP(\mathcal{Y}) \}$, the \textbf{predictive $\cV$-entropy} is defined as:
    \begin{equation}
        H_\cV(Y|\varnothing) = \inf_{f \in \cV} \EEb{ y \sim Y}{-\log{f[\varnothing](y)}} \,.
        \label{v-entropy}
    \end{equation}
    Similarly, the \textbf{conditional $\cV$-entropy} is expressed as:
    \begin{equation}
        H_\cV(Y|X) = \inf_{f \in \cV} \mathbb{E} [- \log f[\xx](y) ] \,.
        \label{cond-v-entropy}
    \end{equation}
    Here, $\log$ quantifies the entropies in nats.
\end{definition}
In essence, $f[\xx]$ and $f[\varnothing]$ generate probability distributions over the labels.
The objective is to identify $f \in \cV$ that maximizes the log-likelihood of the label data, both with (\ref{cond-v-entropy}) and without the input (\ref{v-entropy}).

\begin{definition}
    Consider random variables $X$ and $Y$ with respective sample spaces $\mathcal{X}$ and $\mathcal{Y}$.
    Within the context of a predictive family $\cV$, the \textbf{$\cV$-information} is defined as:
    \begin{equation}
        I_\cV(X \to Y) = H_\cV(Y|\varnothing) - H_\cV(Y|X) \, .
        \label{eq:v-info}
    \end{equation}
\end{definition}

Given the finite nature of the dataset, the estimated $\cV$-information may deviate from its true value.
\citet{xu2020theory} establish PAC bounds for this estimation error, with less complex $\cV$ and larger datasets yielding more precise bounds.
Besides, several key properties of $\cV$-information, enumerated by \citet{xu2020theory}, include:
\begin{itemize}[noitemsep,topsep=0pt]
    \item \emph{Non-Negativity:} $I_\cV(X \to Y) \geq 0$
    \item \emph{Independence:} If $X$ is independent of $Y$, $I_\cV(X \to Y) = I_\cV(Y \to X) =  0$.
    \item \emph{Monotonicity:} For $\cV \subseteq \cU$, $H_\cV(Y|\varnothing) \geq H_\cU(Y|\varnothing)$ and $H_\cV(Y|X) \geq H_\cU(Y|X)$.
\end{itemize}

\subsection{Intuition of $\cV$-information on Distilled Dataset}
Maximizing the $\mathcal{V}$-information $I_\cV(X \to Y)$ for real-world datasets proves intractable, primarily attributed to the inherent disparity between the boundless information sources and the constrained capabilities of observers within the predictive family $\mathcal{V}$.
A promising avenue arises, however, in the form of distilled datasets, wherein information is derived from a finite original full dataset. This ensures the existence of an optimal predictive family $\mathcal{V} \subseteq \Omega$ exemplified by observer models trained on the original full dataset.
Consequently, the realism of the distilled dataset can be precisely assessed by leveraging this (almost) optimal predictive family.

Furthermore, the upper bound of diversity in the distilled dataset can be reliably guaranteed by the finite information (diversity) encapsulated within the original full dataset.
This stands in stark contrast to the challenging task of limiting the diversity inherent in real-world datasets.

\paragraph{Data realism and $\cV$-information.}
Consider an observer (predictive) family $\cV$ capable of mapping image input $X$ to its corresponding label output $Y$.
If we transform the images $X$ into encrypted versions or introduce additional noisy features beyond their natural background noise, predicting $Y$ given $X$ with the same $\cV$ becomes more challenging.

To capture this intuition, a framework termed $\cV$-information~\cite{xu2020theory} generalizes Shannon information~\footnote{
    The conventional approach of using \citet{shannon1948mathematical}'s mutual information $I(X;Y)$ is not suitable in this context.
    This metric remains unchanged after the transformation of $X$, as it permits unbounded computation, including any necessary for the inverse transformation of images.
}, measuring how much information can be extracted from $X$ about $Y$ when constrained to observers in $\cV$, denoted as $I_\cV(X \to Y)$.
When $\cV$ encompasses an infinite set of observers, corresponding to unbounded computation, $\cV$-information reduces to Shannon information.

Likewise, unrealistic output labels for $Y$, such as encrypted or noisy labels, or even simplistic one-hot labels, prove inadequate in representing the precise information contained within images $X$.
This inadequacy leads to diminished predictive accuracy, even when employing robust observers from the set $\cV$.

\paragraph{Data diversity from the perspective of $\cV$-information.}
$\cV$-information $I_\cV(X \to Y)$ serves as a conceptual tool for gauging the interconnected information between images $X$ and labels $Y$.
Consequently, this measurement is inherently influenced by the overall amount of information within both images $X$ and labels $Y$.
However, in the context of natural image datasets like ImageNet-1K~\cite{deng2009imagenet}, the diversity (information entropy) between images $X$ and labels $Y$ is notably imbalanced.
Specifically, the labels $Y$ often encompass considerably less information compared to the images $X$, thereby constraining $\cV$-information $I_\cV(X \to Y)$.

\paragraph{Summary.}
\emph{Enhancing the diversity and realism of both the input $X$ and the output $Y$ in a dataset necessitates maximizing the $\cV$-information $I_\cV(X \to Y)$.}

\subsection{Maximizing $\cV$-information in Practice}

\begin{proof}[Maximizing diversity of distilled data]
    Consider a predictive family $\cV = \{\phi_{\mathrm{h}}, \phi_{\mtheta_{\cT}}\}$ and a distilled dataset $\cS_c =(X_c,Y_c)$ for class $c$ dataset $\cT_c$, we assume:
    \begin{equation}
        \forall \cS_c , \, \exists h\in \cH ,\, \text{s.t.} \, \cS_c = \{(\xx_c,y_c) \mid y_c=h(\xx_c)\} \, ,
    \end{equation}
    where $ \cH = \{h: \cX \to \cY $\}.
    This assumption establishes the upper bound of diversity term for a distilled dataset $\cS_c$, defined by the $\cV$-entropy as follows:
    \begin{equation}
        \begin{split}
            &H_{\cV}(Y_c|\varnothing) \\
            &=\inf_{f \in {\cV}} \mathbb{E} [- \log f[\varnothing](y_c) ] \\
            &=\inf_{f \in {\cV}} \mathbb{E} [- \log f[\varnothing](h(\xx_c)) ] \\
            &\leq\inf_{f \in {\cV}} \mathbb{E} [- \log f[\varnothing](\xx_c) ] \\
            &=H_{\cV}(X_c|\varnothing) \, .
        \end{split}
    \end{equation}
    Given $\cT_c=(\hat{X}_{c},\hat{Y}_{c})$, where $(\hat{X}_{c},\hat{Y}_{c}) := \{ (\hat{\xx}, \hat{y}) \vert (\hat{\xx}, \hat{y}) \in \cT, \hat{y} \!=\! c \}$, we have:
    \begin{equation}
        \begin{split}
            &H_{\cV}(\cT_c|\varnothing) \\
            &=H_{\cV}((\hat{X}_{c},\hat{Y}_{c})|\varnothing) \\
            &=\inf_{f \in {\cV}} \mathbb{E} [- \log f[\varnothing](\hat{\xx}_{c},\hat{y}_{c}) ] \\
            &=\inf_{f \in {\cV}} \mathbb{E} [- \log f[\varnothing](\hat{\xx}_{c},{c}) ] \\
            &=\inf_{f \in {\cV}} \mathbb{E} [- \log f[\varnothing](\hat{\xx}_{c}) ] \\
            &\geq\inf_{f \in {\cV}} \mathbb{E} [- \log f[\varnothing](\xx_c) ] \\
            &\geq H_{\cV}(Y_c|\varnothing)\, .
        \end{split}
    \end{equation}
    Consequently, the above theoretical analysis can be extended to the entire dataset $\cT$ and obtain that:
    $H_{\cV}(Y|\varnothing) \leq H_{\cV}(X|\varnothing) \leq H_{\cV}(\cS|\varnothing) \leq H_{\cV}(\cT|\varnothing)=C$,
    where $C$ is a constant for a certain $\cT$.
    Thus, we obtain:
    \begin{equation}
        {H_{\cV}(Y|\varnothing)}\propto{H_{\cV}(Y|\varnothing)/H_{\cV}(\cT|\varnothing)} \leq 1\, .
    \end{equation}
    If we maximize the diversity term ${H_{\cV}(Y|\varnothing)}$, then the ratio ${H_{\cV}(Y|\varnothing)/H_{\cV}(\cT|\varnothing)}=1$ and ${H_{\cV}(\cS|\varnothing)=H_{\cV}(\cT|\varnothing)} $.
\end{proof}

\begin{proof}[Maximizing realism of distilled data]
    Given a predictive family $\cV = \{\phi_{\mathrm{h}}, \phi_{\mtheta_{\cT}}\}$ and a distilled dataset $\cS =(X,Y)$, our objective is to minimize the realism term defined by the conditional $\cV$-entropy:
    \begin{equation}
        \begin{split}
            &H_{\cV}(Y|X) \\
            &=\inf_{f \in {\cV}} \mathbb{E} [- \log f[\xx](y) ] \\
            &\leq\mathbb{E} [- \log \phi_{\mathrm{h}}[\xx](y) ] + \mathbb{E} [- \log \phi_{\mtheta_{\cT}}[\xx](y) ] \, .
        \end{split}
    \end{equation}
    To estimate the density value $f[\xx](y)$, we adopt the approach proposed by \citet{oord2018representation}:
    \begin{equation}
        f[\xx](y) = \frac{\exp(-\ell(f(\xx), y))}{\E_{y^\prime \in Y}{[\exp(-\ell(f(\xx), y^\prime))]}}  \,,
        \label{eq:density}
    \end{equation}
    leading to:
    \begin{equation}
        \begin{split}
            &H_{\cV}(Y|X) \\
            &\leq\mathbb{E} [- \log \frac{\exp(-\ell(\phi_{\mathrm{h}}(\xx), y))}{\E_{y^\prime \in Y}{[\exp(-\ell(\phi_{\mathrm{h}}(\xx), y^\prime))]}} ] \\
            &\quad + \mathbb{E} [- \log \frac{\exp(-\ell(\phi_{\mtheta_{\cT}}(\xx), y))}{\E_{y^\prime \in Y}{[\exp(-\ell(\phi_{\mtheta_{\cT}}(\xx), y^\prime))]}} ] \, .
        \end{split}
    \end{equation}
    Assuming the function $\ell(\cdot)$ is symmetric, i.e.,
    \begin{equation}
        \forall z_1,z_2 , \, \text{s.t.} \, \ell(z_1,z_2)=\ell(z_2,z_1) \, ,
    \end{equation}
    thus, we derive an alternative objective for minimization:
    \begin{equation}
        \begin{split}
            &H_{\cV}(Y|X) \\
            &\propto\mathbb{E} [- \log \frac{\exp(-\ell(\phi_{\mathrm{h}}(\xx), \phi_{\mtheta_{\cT}}(\xx)))}{\E_{\xx \in X}{[\exp(-\ell(\phi_{\mathrm{h}}(\xx), \phi_{\mtheta_{\cT}}(\xx)))]}} ] \\
            &\quad + \mathbb{E} [- \log \frac{\exp(-\ell(\phi_{\mtheta_{\cT}}(\xx), y))}{\E_{y^\prime \in Y}{[\exp(-\ell(\phi_{\mtheta_{\cT}}(\xx), y^\prime))]}} ] \\
            &\propto\mathbb{E} [- \log \exp(-\ell(\phi_{\mathrm{h}}(\xx), \phi_{\mtheta_{\cT}}(\xx))) ] \\
            &\quad + \mathbb{E} [- \log \exp(-\ell(\phi_{\mtheta_{\cT}}(\xx), y)) ] \\
            &=\mathbb{E} [\ell(\phi_{\mathrm{h}}(\xx), \phi_{\mtheta_{\cT}}(\xx))+ \ell(\phi_{\mtheta_{\cT}}(\xx), y)] \, .
        \end{split}
    \end{equation}
    This analysis underpins our strategy to enhance the realism of distilled data by minimizing $H_{\cV}(Y|X)$, we focus on samples $\xx$ that minimize $\ell(\phi_{\mathrm{h}}(\xx), \phi_{\mtheta_{\cT}}(\xx))$ and set $y=\phi_{\mtheta_{\cT}}(\xx)$.
\end{proof}

\section{Detailed Implementation}
\label{sec:imple_details}

\subsection{Pre-training Observer Models}
Following prior studies~\cite{yin2023squeeze,zhao2023improved,cazenavette2022dataset,guo2023towards}, we employ pre-trained observer models to distill the dataset, as illustrated in Table~\ref{tb:main}: 1) ResNet-18 for ImageNet-10, ImageNette, ImageWoof, ImageNet-100, ImageNet-1K; 2) modified ResNet-18 for CIFAR-10, CIFAR-100 and Tiny-ImageNet; 3) ConvNet-3 for CIFAR-10, CIFAR-100; 4) ConvNet-4 for Tiny-ImageNet; 5) ConvNet-5 for ImageWoof, ImageNette; 6) ConvNet-6 for ImageNet-100.

\subsection{Implementing \algopt algorithm.}
To gain an intuitive understanding the Algorithm~\ref{alg:framework} of our proposed~\algopt, we expound on the implementation details in this section.
Given a comprehensive real dataset $\cT$, such as ImageNet-1K~\cite{deng2009imagenet}, we define three tasks involving distilling this dataset into smaller datasets with distinct \IPC values, specifically, $\IPC=50$, $10$, and $1$.
Remarkably, our \algopt demonstrates the capability to encompass multisize distilled datasets through a single distillation process, effectively handling those with $\IPC=50$, $10$, and $1$.

\paragraph{Extracting key patches.}

For each class set $\cT_c$ we uniformly pre-select a subset contains $300$ images denoted as $\cT_c^\prime = \{\hat{\xx}_i\}_{i=1}^{300}$.
Each pre-selected image $\hat{\xx}_i$ undergoes random cropping into $K=5$ patches\footnote{
    We empirically set $K=5$, although smaller values, such as $K=1$, can be chosen for expedited implementation of our algorithm \algopt.
}.
These patches are represented as $\{\xi_{i,k}\}_{k=1}^{K=5}$, and the realism score $s_{i,k} = -\ell({\phi_{\mtheta_{\cT}}}(\xi_{i,k}), y_i)$ is calculated for each patch $\xi_{i,k}$, resulting in a set of scores $\{s_{i,k}\}_{k=1}^{K=5}$.
Subsequently, the key patch $\xi_{i,\star}$ with the highest realism score $s_{i,\star}$ is selected to represent the corresponding image $\xx_i$.
This process yields a key patch set with scores $\{\xi_{i,\star}, s_{i,\star}\}_{i=1}^{300}$, which is stored for future use.

\paragraph{Capturing class information.}
We prioritize key patches, denoted as $\{\xi_{i,\star}\}_{i=1}^{300}$, based on their associated scores $\{s_{i,\star}\}_{i=1}^{300}$ to construct a well-ordered set $\{\xi_{j,\star}\}_{j=1}^{300}$.
In addressing the initial task of synthesizing a refined dataset with $\IPC = 50$, we strategically choose the top-$(200 = \IPC \times N)$ key patches from the set, denoted as $\{\xi_{j,\star}\}_{j=1}^{200}$.
Likewise, for the two subsequent tasks, characterized by $\IPC = 10$ and $\IPC = 1$, we iteratively refine the selection by opting for the top-$40$ and top-$4$ key patches, denoted as $\{\xi_{j,\star}\}_{j=1}^{40}$ and $\{\xi_{j,\star}\}_{j=1}^{4}$, respectively.

\paragraph{Images reconstruction.}
To construct the ultimate image ${\xx}_j$, we systematically draw $N=4$ distinct patches $\{\xi_{j,\star}\}_{j=1}^{N=4}$ without replacement and concatenate them.
This procedure is iterated times to generate the ultimate distilled image set $\{{\xx}_j\}_{j=1}^{\IPC}$.

\paragraph{Labels reconstruction.}

In accordance with the methodology presented in SRe$^2$L~\cite{yin2023squeeze}, we undertake the process of relabeling the distilled images through the generation and storage of region-level soft labels, denoted as ${y}_j$, employing Fast Knowledge Distillation~\cite{shen2022fast}.
To achieve this, for each distilled image ${\xx}_j$, we perform random cropping into several patches, concurrently documenting their coordinates on the image ${\xx}_j$.
Subsequently, soft labels ${y}_{j,m}$ are generated and stored for each $m$-th patch, ultimately culminating in the aggregation of these labels to form the comprehensive ${y}_j$.

\subsection{Training on Distilled Dataset}
Following prior investigations~\cite{yin2023squeeze,cui2022dc,yu2023dataset}, we employ data-augmentation techniques, namely RandomCropResize~\cite{wong2016understanding} and CutMix~\cite{yun2019cutmix}.
Further elucidation is available in our publicly accessible code repository at \textit{https://to-be-released}.

\begin{table*}[ht]
    \centering
    \setlength\tabcolsep{5.5pt}
    \scalebox{0.84}{
        \begin{tabular}{@{}c|ccccc|ccc@{}}
            \toprule[2pt]
                                             & \multicolumn{5}{c|}{RDED (Ours)} & \multicolumn{3}{c}{SRe$^2$L}                                                                                                                 \\
            Verifier\textbackslash{}Observer & ResNet-18                        & EfficientNet-B0              & MobileNet-V2            & VGG-11         & Swin-V2-Tiny   & ResNet-18      & EfficientNet-B0 & MobileNet-V2   \\ \midrule
            ResNet-18                        & \textbf{42.3 $\pm$ 0.6}          & \textbf{31.0 $\pm$ 0.1}      & \textbf{40.4 $\pm$ 0.1} & 36.6 $\pm$ 0.1 & 17.2 $\pm$ 0.2 & 21.7 $\pm$ 0.6 & 11.7 $\pm$ 0.2  & 15.4 $\pm$ 0.2 \\
            EfficientNet-B0                  & \textbf{42.8 $\pm$ 0.5}          & \textbf{33.3 $\pm$ 0.9}      & \textbf{43.6 $\pm$ 0.2} & 35.8 $\pm$ 0.5 & 14.8 $\pm$ 0.1 & 25.2 $\pm$ 0.2 & 11.4 $\pm$ 2.5  & 20.5 $\pm$ 0.2 \\
            MobileNet-V2                     & \textbf{34.4 $\pm$ 0.2}          & \textbf{24.1 $\pm$ 0.8}      & \textbf{33.8 $\pm$ 0.6} & 28.7 $\pm$ 0.2 & 11.8 $\pm$ 0.3 & 19.7 $\pm$ 0.1 & 9.8 $\pm$ 0.4   & 10.2 $\pm$ 2.6 \\
            VGG-11                           & \textbf{22.7 $\pm$ 0.1}          & \textbf{16.5 $\pm$ 0.8}      & \textbf{21.6 $\pm$ 0.2} & 23.5 $\pm$ 0.3 & 7.8 $\pm$ 0.1  & 16.5 $\pm$ 0.1 & 9.3 $\pm$ 0.1   & 10.6 $\pm$ 0.1 \\
            Swin-V2-Tiny                     & \textbf{17.8 $\pm$ 0.1}          & \textbf{19.7 $\pm$ 0.3}      & \textbf{18.1 $\pm$ 0.2} & 15.3 $\pm$ 0.4 & 12.1 $\pm$ 0.2 & 9.6 $\pm$ 0.3  & 10.2 $\pm$ 0.1  & 7.4 $\pm$ 0.1  \\ \bottomrule[2pt]
        \end{tabular}}
    \vspace{-1em}
    \caption{
        \textbf{Evaluating ImageNet-1K top-1 accuracy on cross-architecture generalization.}
        Distill dataset with VGG-11~\cite{simonyan2014very}, Swin-V2-Tiny~\cite{liu2022swin}, ResNet-18 \cite{he2016deep}, EfficientNet-B0 \cite{tan2019efficientnet}, MobileNet-V2 \cite{sandler2018mobilenetv2}, and then versus transfer to other each other architecture.
    }
    \label{tb:more_crossarch}
\end{table*}

\section{Experiment}
\label{sec:aux_exp}
In this section, unless otherwise specified, we adopt ResNet-18 as the default neural network backbone for both the distillation process and subsequent evaluation.
The parameters $\IPC = 10$ and pre-selected subset size $\abs{\cT_c^\prime} = 300$ are consistently applied.
For high-resolution datasets, we set the number of patches $N=4$ within one distilled image, while for datasets with a resolution lower than $64 \times 64$, we use $N=1$.
All settings are consistent with those in Section~\ref{sec:exp}.

\subsection{Multisize Dataset Distillation}
In their recent work, \citet{he2024multisize} introduced Multisize Dataset Condensation (MDC), a novel approach that consolidates multiple condensation processes into a unified procedure.
This innovative method produces datasets with varying sizes, offering dual advantages:
\begin{itemize}
    \item DC eliminates the necessity for extra condensation processes when distilling multiple datasets with varying $\IPC$.
    \item It facilitates a reduction in storage requirements by reusing condensed images.
\end{itemize}
Remarkably, our proposed \algopt, also exhibits a mechanism that enables the synthesis of distilled datasets with adaptable $\IPC$ without incurring additional computational overhead (c.f. Section~\ref{sec:imple_details}).
For a comprehensive comparison, the superior performance of our \algopt over MDC on larger distilled datasets is demonstrated in Table~\ref{tb:comparison_mdc}.

\begin{table}[ht]
    \centering
    \setlength\tabcolsep{6.1pt}
    \begin{tabular}{@{}c|ccc|ccc@{}}
        \toprule[2pt]
        \multicolumn{1}{c|}{}        & \multicolumn{3}{c|}{CIFAR-10} & \multicolumn{3}{c}{CIFAR-100}                                                                 \\
        Method \textbackslash{} \IPC & 1                             & 10                            & 50            & 1             & 10            & 50            \\ \midrule
        MDC                          & \textbf{47.8}                 & \textbf{62.6}                 & \textbf{74.6} & \textbf{26.3} & 41.4          & 53.7          \\
        Ours                         & 23.5                          & 50.2                          & 68.4          & 19.6          & \textbf{50.2} & \textbf{57.0} \\ \bottomrule[2pt]
    \end{tabular}
    \vspace{-1em}
    \caption{
        \textbf{Comparison with Multisize Dataset Condensation}.
        The top-1 validation accuracy is evaluated when both MDC and our \algopt are targeting at distilling dataset with $\IPC = 50$. The other two distilled datasets with $\IPC = 10$ and $\IPC = 1$ are subsets from the one with $\IPC = 50$.
        The neural network backbone used for distillation and evaluation is Conv-3.
    }
    \label{tb:comparison_mdc}
\end{table}

\subsection{CoreSet Selection Baselines}

In our investigation, we assess the top-1 validation accuracy resulting from the application of three CoreSet selection strategies for dataset distillation: 1) Random; 2) Herding~\cite{welling2009herding}; 3) K-Means~\cite{forgy1965cluster}.
The outcomes, as depicted in Table~\ref{tb:pruning_baselines}, indicate catastrophically poor performance when employing these selection methods directly in the context of dataset distillation.

\begin{table}[ht]
    \centering
    \setlength\tabcolsep{7.7pt}
    \begin{tabular}{@{}c|ccc@{}}
        \toprule[2pt]
        Dataset       & Random                  & Herding                 & K-Means                 \\ \midrule
        ImageNet-10   & \textbf{36.7 $\pm$ 0.1} & 33.8 $\pm$ 0.4          & 36.5 $\pm$ 0.3          \\
        ImageNet-100  & 10.8 $\pm$ 0.2          & 12.6 $\pm$ 0.1          & \textbf{13.5 $\pm$ 0.4} \\
        ImageNet-1K   & 4.4 $\pm$ 0.1           & \textbf{5.8 $\pm$ 0.1}  & 5.5 $\pm$ 0.1           \\
        Tiny-ImageNet & 7.5 $\pm$ 0.1           & \textbf{9.0 $\pm$ 0.3}  & 8.9 $\pm$ 0.2           \\
        CIFAR-100     & 10.9 $\pm$ 0.1          & \textbf{13.3 $\pm$ 0.3} & 12.9 $\pm$ 0.1          \\
        CIFAR-10      & 25.1 $\pm$ 0.5          & \textbf{28.4 $\pm$ 0.1} & 27.7 $\pm$ 0.2          \\ \bottomrule[2pt]
    \end{tabular}
    \vspace{-1em}
    \caption{
        \textbf{Comparison of different CoreSet selection-based dataset distillation baselines}.
        Experiments are carried out to evaluate three widely used coreset selection methods.
    }
    \vspace{-1em}
    \label{tb:pruning_baselines}
\end{table}

\subsection{Cross-architecture Generalization}
We expanded our experimental evaluations by incorporating various neural network architectures that lack batch normalization~\cite{ioffe2015batch,yin2023squeeze}.
This extension aims to thoroughly assess the cross-architecture generalization capabilities of our proposed \algopt.
The results presented in Table~\ref{tb:more_crossarch} unequivocally demonstrate the superior performance of \algopt in comparison to the SOTA method SRe$^2$L.
Notably, our algorithm exhibits remarkable effectiveness even in scenarios characterized by substantial architectural disparities, such as knowledge transfer from ResNet-18 to Swin-V2-Tiny.

\subsection{Detailed Ablation Study}
\label{sec:aux_ablationstudy}
In addition to the experiments detailed in Section~\ref{sec:ablationstudy}, we conduct a more comprehensive ablation study, delving into the various approaches and hyperparameters employed in our proposed \algopt.

\paragraph{On the impact of $|\cT_c^\prime|$ and $N$.}

\begin{figure}[ht]
    \begin{minipage}{0.20\textwidth}
        \centering
        \includegraphics[width=1.25\linewidth]{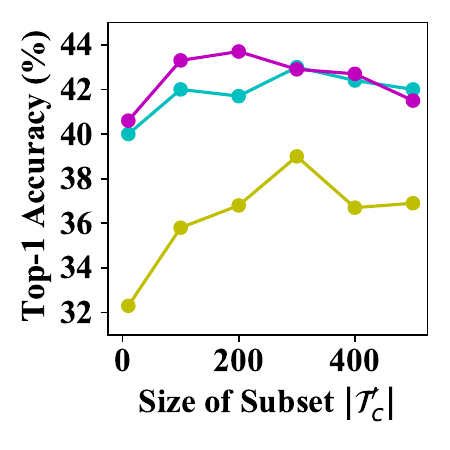}
    \end{minipage}
    \hspace{0.6cm}
    \begin{minipage}{0.20\textwidth}
        \centering
        \includegraphics[width=1.15\linewidth]{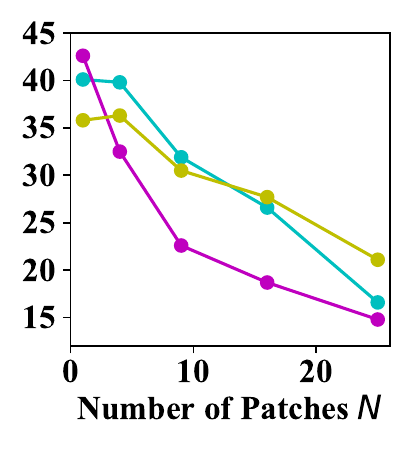}
    \end{minipage}
    \vspace{-1.5em}
    \caption{
        \textbf{Ablation study on $\abs{\cT_c^\prime}$ and $N$}, i.e., the pre-selected subset size $\cT_c^\prime$ (left), and the number of patches $N$ within each distilled image (right).
        The lemon \textcolor{lemon}{$\bullet$}, purple \textcolor{purple}{$\bullet$}, and turquoise \textcolor{turquoise}{$\bullet$} denote CIFAR-10, CIFAR-100, and Tiny-ImageNet respectively.
        \looseness=-1
    }
    \label{fig:elements_withlow}
\end{figure}

To assess the influence of the pre-selected subset size $|\cT_c^\prime|$ and the number of patches within each distilled image $N$, our experiments are extended to lower-resolution datasets, namely Tiny-ImageNet, CIFAR-10, and CIFAR-100.
Figure~\ref{fig:elements_withlow} illustrates that the configurations with $|\cT_c^\prime| = 300$ and $N = 1$ are suitable for low-resolution datasets.

\begin{table*}[ht]
    \centering
    \setlength\tabcolsep{21.7pt}
    \begin{tabular}{@{}c|ccccc@{}}
        \toprule[2pt]
        Dataset       & Original       & +EKP           & +CCI           & +IR            & +LR                     \\ \midrule
        ImageNet-10   & 30.6 $\pm$ 0.4 & 34.5 $\pm$ 1.1 & 39.6 $\pm$ 1.6 & 49.9 $\pm$ 1.5 & \textbf{54.3 $\pm$ 2.7} \\
        ImageNet-100  & 8.2 $\pm$ 0.2  & 9.8 $\pm$ 0.1  & 15.0 $\pm$ 0.5 & 24.1 $\pm$ 0.1 & \textbf{35.9 $\pm$ 0.1} \\
        ImageNet-1K   & 3.2 $\pm$ 0.1  & 3.8 $\pm$ 0.1  & 7.2 $\pm$ 0.3  & 15.2 $\pm$ 0.1 & \textbf{42.1 $\pm$ 0.1} \\
        Tiny-ImageNet & 6.9 $\pm$ 0.1  & 8.8 $\pm$ 0.1  & 15.7 $\pm$ 0.2 & -              & \textbf{41.9 $\pm$ 0.2} \\
        CIFAR-100     & 11.8 $\pm$ 0.1 & 13.2 $\pm$ 0.3 & 18.6 $\pm$ 0.3 & -              & \textbf{42.6 $\pm$ 0.1} \\
        CIFAR-10      & 27.7 $\pm$ 0.6 & 26.8 $\pm$ 0.2 & 27.8 $\pm$ 0.5 & -              & \textbf{35.8 $\pm$ 0.0} \\ \bottomrule[2pt]
    \end{tabular}
    \vspace{-1em}
    \caption{
        \textbf{Effectiveness of accumulated techniques in \algopt}.
        The validation accuracy undergoes a gradual evolution as we sequentially apply the four techniques in our \algopt.
        Entries marked with ``-'' are absent because of the $N = 1$ setting for low-resolution datasets, rendering the Images Reconstruction (IR) step impractical.
    }
    \label{tb:techniques}
\end{table*}

\paragraph{Effectiveness of each technique in \algopt.}
To validate the effectiveness of all four components within our \algopt, we conduct additional ablation studies for each of them, namely, Extracting Key Patches (EKP), Capturing Class Information (CCI), Images Reconstruction (IR), and Labels Reconstruction (LR), corresponding to the techniques outlined in Sections~\ref{sec:infoextract} and~\ref{sec:inforecon}.
Table~\ref{tb:techniques} illustrates that all four techniques employed in \algopt are essential for achieving the remarkable final performance.
Furthermore, a plausible hypothesis suggests that LR plays a crucial role in generating more informative (diverse) and aligned (realistic) labels for distilled images, thereby significantly enhancing performance.

\paragraph{Effectiveness of selecting patches through realism socre.}

\begin{table}[ht]
    \centering
    \setlength\tabcolsep{2.7pt}
    \scalebox{0.92}{
        \begin{tabular}{@{}c|cccc@{}}
            \toprule[2pt]
            Dataset       & Random         & Herding                 & K-Means                 & Realism                 \\ \midrule
            ImageNet-10   & 44.7 $\pm$ 2.5 & 47.9 $\pm$ 0.3          & 49.3 $\pm$ 1.1          & \textbf{53.3 $\pm$ 0.1} \\
            ImageNet-100  & 29.8 $\pm$ 0.7 & 29.7 $\pm$ 0.5          & 28.9 $\pm$ 0.1          & \textbf{36.0 $\pm$ 0.3} \\
            ImageNet-1K   & 37.9 $\pm$ 0.5 & 38.4 $\pm$ 0.1          & 38.2 $\pm$ 0.1          & \textbf{42.0 $\pm$ 0.1} \\
            Tiny-ImageNet & 40.2 $\pm$ 0.0 & 41.1 $\pm$ 0.1          & 40.1 $\pm$ 0.1          & \textbf{41.9 $\pm$ 0.2} \\
            CIFAR-100     & 41.4 $\pm$ 0.5 & \textbf{42.6 $\pm$ 0.1} & 41.8 $\pm$ 0.1          & \textbf{42.6 $\pm$ 0.1} \\
            CIFAR-10      & 34.3 $\pm$ 0.1 & 35.5 $\pm$ 0.6          & \textbf{37.9 $\pm$ 0.3} & 35.8 $\pm$ 0.1          \\ \bottomrule[2pt]
        \end{tabular}
    }
    \vspace{-1em}
    \caption{
        \textbf{Comparison of different patch selection strategies in \algopt}.
        Experiments are conducted to compare our proposed realism-score-based data selection strategy over three widely used coreset selection methods.
    }
    \vspace{-1em}
    \label{tb:pruning}
\end{table}

Table~\ref{tb:pruning} demonstrates that our realism-score-based selection method, specifically the Capturing Class Information (CCI) technique outlined in Algorithm~\ref{alg:framework}, consistently outperforms alternative approaches, except for CIFAR-10.
A plausible inference is that the selection of more realistic images contributes to the observer model's ability to reconstruct correspondingly realistic labels (cf. Section~\ref{sec:inforecon}), thereby optimizing our objective \eqref{eq:principle}.

\end{document}